\RequirePackage{fix-cm}
\documentclass[twocolumn]{svjour3}          
\smartqed  
\usepackage{graphicx}
\usepackage{color}
\usepackage{natbib}
\usepackage{amsmath,amssymb} 
\usepackage{multirow}
\usepackage{tabu}
\journalname{International Journal of Computer Vision}
\begin{document}

\title{Multi-level Motion Attention for Human Motion Prediction
}


\author{Wei Mao         \and
        Miaomiao Liu  \and 
        Mathieu Salzmann \and
        Hongdong Li
}


\institute{
            Wei Mao \at
            Australian National University\\
            \email{wei.mao@anu.edu.au}
            \and
            Miaomiao Liu \at
            Australian National University \\
            \email{miaomiao.liu@anu.edu.au}
            \and
            Mathieu Salzmann \at
            EPFL--CVLab \& ClearSpace, Switzerland\\
            \email{mathieu.salzmann@epfl.ch}
            \and
            Hongdong Li \at
            Australian National University\\
            \email{hongdong.li@anu.edu.au}
}

\date{Received: 14 September 2020 / Accepted: 24 May 2021}

\maketitle

\begin{abstract}
Human motion prediction aims to forecast future human poses given a historical motion. Whether based on recurrent or feed-forward neural networks, existing learning based methods fail to model the observation that human motion tends to repeat itself, even for complex sports actions and cooking activities. Here, we introduce an attention based feed-forward network that explicitly leverages this observation. In particular, instead of modeling frame-wise attention via pose similarity, we propose to extract \emph{motion attention} to capture the similarity between the current motion context and the historical motion sub-sequences. In this context, we study the use of different types of attention, computed at joint, body part, and full pose levels.
Aggregating the relevant past motions and processing the result with a graph convolutional network allows us to effectively exploit motion patterns from the long-term history to predict the future poses. Our experiments on Human3.6M, AMASS and 3DPW validate the benefits of our approach for both periodical and non-periodical actions. Thanks to our attention model, it yields state-of-the-art results on all three datasets. Our code is available at
\noindent https://github.com/wei-mao-2019/HisRepItself.

\end{abstract}
\section{Introduction}\label{sec:intro}
Human motion prediction consists of forecasting the future poses of a person given a history of their previous motion. Predicting human motion can be highly beneficial for tasks such as human tracking~\citep{gong2011multi}, human-robot interaction~\citep{koppula2013anticipating}, and human motion generation for computer graphics~\citep{2012-ccclde,kovar2008motion,sidenbladh2002implicit}.

\begin{figure*}[!ht]
    \centering
      \includegraphics[width=\linewidth]{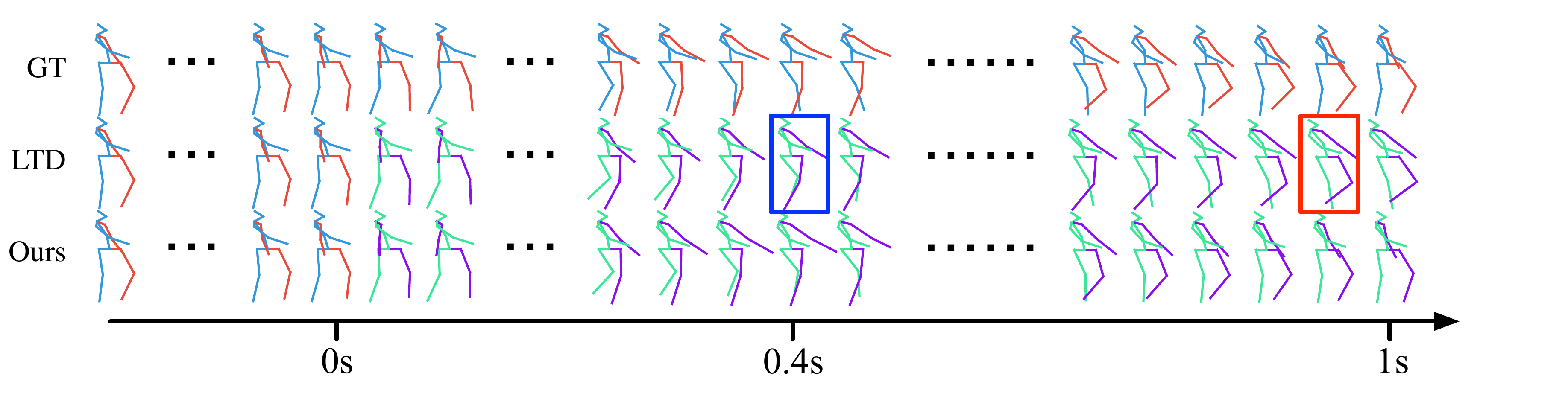}
      \caption{{\bf Human motion prediction} aims to forecast future human poses ($>0s$) given past ones. From top to bottom, we show the ground-truth pose sequence, the predictions of LTD~\citep{mao2019learning} and those of our approach. The frames where LTD~\citep{mao2019learning} yields larger errors on the arms and legs are highlighted with a blue and red box, respectively. Note that our results better match the ground truth than those of LTD~\citep{mao2019learning}.}
      \label{fig:intro}
\end{figure*}
Traditional methods, such as hidden Markov models~\citep{brand2000style} and Gaussian Process Dynamical Models~\citep{wang2008gaussian}, have proven effective for simple motions, such as walking and golf swings. However, they are typically outperformed by deep learning ones on more complex motions.
The most common trend in modeling the sequential data that constitutes human motion consists of using Recurrent Neural Networks (RNNs)\citep{Martinez_2017_CVPR,fragkiadaki2015recurrent,JainZSS16}. However, as discussed in~\citep{LiZLL18}, in the mid- to long-term horizon, RNNs tend to generate static poses because they struggle to keep track of long-term history. To tackle this problem, existing works~\citep{LiZLL18,gui2018adversarial} either rely on Generative Adversarial Networks (GANs), which are  notoriously hard to train~\citep{ArjovskyB17}, or introduce an additional \emph{long-term encoder} to represent information from the further past~\citep{LiZLL18}. Unfortunately, such an encoder treats the entire motion history equally, thus not allowing the model to put more emphasis on the parts of the past motion that better reflect the context of the current motion.

In this paper, by contrast, we introduce an attention-based motion prediction approach that effectively exploits historical information by dynamically adapting its focus on the previous motions to the current context. Our method is motivated by the observation that humans tend to repeat their motion, not only in short periodical activities, such as walking, but also in more complex actions occurring across longer time periods, such as sports and cooking activities~\citep{runia2018real,li2018structure}. Therefore, we aim to find the relevant historical information to predict future motion.

While~\cite{Tang_2018} have attempted to leverage attention for motion prediction, 
they achieved this in a frame-wise manner,
by comparing the human pose from the last observable frame with each one in the historical sequence. As such, this approach fails to reflect the motion direction and is affected by the fact that similar poses may appear in completely different motions. For instance, in most Human3.6M activities, the actor will at some point be standing with their arm resting along their body. To overcome this, we therefore propose to model \emph{motion attention}, and thus compare the last visible sub-sequence with a history of motion sub-sequences.

For periodical motions, such as walking and jogging, humans tend to repeat their full-body motion across long time horizons. However, for non-periodical motions, such as discussion and cooking, motion repetitiveness rather happens at the level of body parts. To model this, we  explore the use of motion attention at three different levels: full pose, body parts, and individual joints. We observe that, as they capture different kinds of motion repetitiveness, the effectiveness of each of these different levels of motion attention varies across different activities and sequences. To handle this, we therefore introduce a fusion model that combines different attention levels and focuses on the attention level best-suited for the current motion context.

When dealing with a time-related problem such as motion prediction, the question of how to encode the temporal information naturally arises. 
The most common trend consists of using Recurrent Neural Networks (RNNs)~\citep{fragkiadaki2015recurrent,JainZSS16,Martinez_2017_CVPR,gui2018adversarial}. However, as argued in~\citep{gui2018adversarial,LiZLL18}, RNNs for motion prediction suffer from error accumulation and discontinuities between the last observed frame and the first predicted one. As an alternative, convolutions  across  time  on  the  observed  poses are used in~\citep{Butepage_2017_CVPR,LiZLL18}. The temporal dependencies that such an approach can encode, however, strongly depend on the size of the convolutional filters. To remove such a dependency, here, we introduce a drastically different approach to modeling temporal information for motion prediction. Inspired by ideas from the nonrigid structure-from-motion literature~\citep{akhter2009nonrigid}, we propose to represent human motion in trajectory space instead of pose space, and thus adopt the Discrete Cosine Transform (DCT) to encode temporal information.

Another question that arises when working with human poses is how to encode the spatial dependencies among the joints. In~\cite{Butepage_2017_CVPR}, this was achieved by exploiting the human skeleton, and in~\cite{LiZLL18} by defining a relatively large spatial filter size. While the former does not allow one to model dependencies across different limbs, such as left-right symmetries, the dependencies encoded by the latter again depend on the size of the filters. In this paper, we propose to overcome these two issues by exploiting graph convolutions~\citep{kipf2016semi}. However, instead of using a pre-defined, sparse graph as in~\citep{kipf2016semi}, we introduce an approach to learning the graph connectivity. This strategy allows the network to capture joint dependencies that are neither restricted to the kinematic tree, nor arbitrarily defined by a convolutional kernel size.

Altogether, our overall framework represents each sub-sequence in trajectory space using the discrete cosine transform (DCT). We then exploit our motion attention at different levels as weights to aggregate the entire DCT-encoded motion history into a future motion estimate. This estimate is combined with the latest observed motion, and the result acts as input to a graph convolutional network (GCN), which lets us better encode spatial dependencies between the different joints. As evidenced by our experiments on Human3.6M~\citep{h36m_pami}, AMASS~\citep{AMASS:2019}, and 3DPW~\citep{vonMarcard2018}, and illustrated in Fig.~\ref{fig:intro}, our motion attention-based approach consistently outperforms the state of the art on short-term and long-term motion prediction by training a single unified model for both settings. This contrasts with our previous, state-of-the-art LTD model~\citep{mao2019learning},  which requires training different models for different settings to achieve its best performance.
Furthermore, we demonstrate that our approach can effectively leverage motion repetitiveness in even longer sequences.

Our contributions can be summarized as follows. (i) We introduce an attention-based model that exploits motions instead of static frames to better leverage historical information for motion prediction; (ii) Our motion attention allows us to train a unified model for both short-term and long-term prediction; (iii) Our approach can effectively make use of motion repetitiveness in long-term history; (iv) It yields state-of-the-art results and generalizes better than existing methods across datasets and actions.

This article extends our previous works~\citep{mao2019learning,mao2020history} in the following ways:
\begin{itemize}
    \item Instead of modeling attention on the full body only, as in~\citep{mao2020history}, we study the use of attention at three different  levels: full body, body parts, and individual joints. Our experiments evidence that different activities or sequences benefit from different levels of attention.
    \item We introduce a fusion module that combines our multi-level attention mechanisms to achieve better performance than the full body pose-level attention model we proposed in~\citep{mao2020history}.
\end{itemize}
\section{Related Work}
\noindent{{\bf RNN-based human motion prediction.}} ~RNNs have proven highly successful in sequence-to-sequence prediction tasks~\citep{sutskever2011generating,kiros2015skip}. As such, they have been widely employed for human motion prediction~\citep{fragkiadaki2015recurrent,JainZSS16,Martinez_2017_CVPR,gopalakrishnan2019neural}. 
For instance, \cite{fragkiadaki2015recurrent} proposed an Encoder Recurrent Decoder~(ERD) model that incorporates a non-linear multi-layer feed-forward network to encode and decode motion before and after recurrent layers. To avoid error accumulation, curriculum learning was adopted during training. \cite{JainZSS16} introduced a Structural-RNN model relying on a manually-designed spatio-temporal graph to encode motion history. 
The fixed structure of this graph, however, restricts the flexibility of this approach at modeling long-range spatial relationships between different limbs. To improve motion estimation, \cite{Martinez_2017_CVPR} proposed a residual-based model that predicts velocities instead of poses. Furthermore, it was shown in this work that a simple zero-velocity baseline, i.e., constantly predicting the last observed pose, led to better performance than~\citep{fragkiadaki2015recurrent,JainZSS16}. While this led to better performance than the previous pose-based methods, the predictions produced by the RNN still suffer from discontinuities between the observed poses and predicted ones.
To overcome this, \cite{gui2018adversarial} proposed to adopt adversarial training to generate smooth sequences. \cite{hernandez2019human} treat human motion prediction as a tensor inpainting problem and exploit a generative adversarial network for long-term prediction. While this approach further improves performance, the use of an adversarial classifier notoriously complicates training~\citep{ArjovskyB17}, making it challenging to deploy on new datasets. 
\begin{figure*}[!t]
    \centering
      \includegraphics[width=\textwidth]{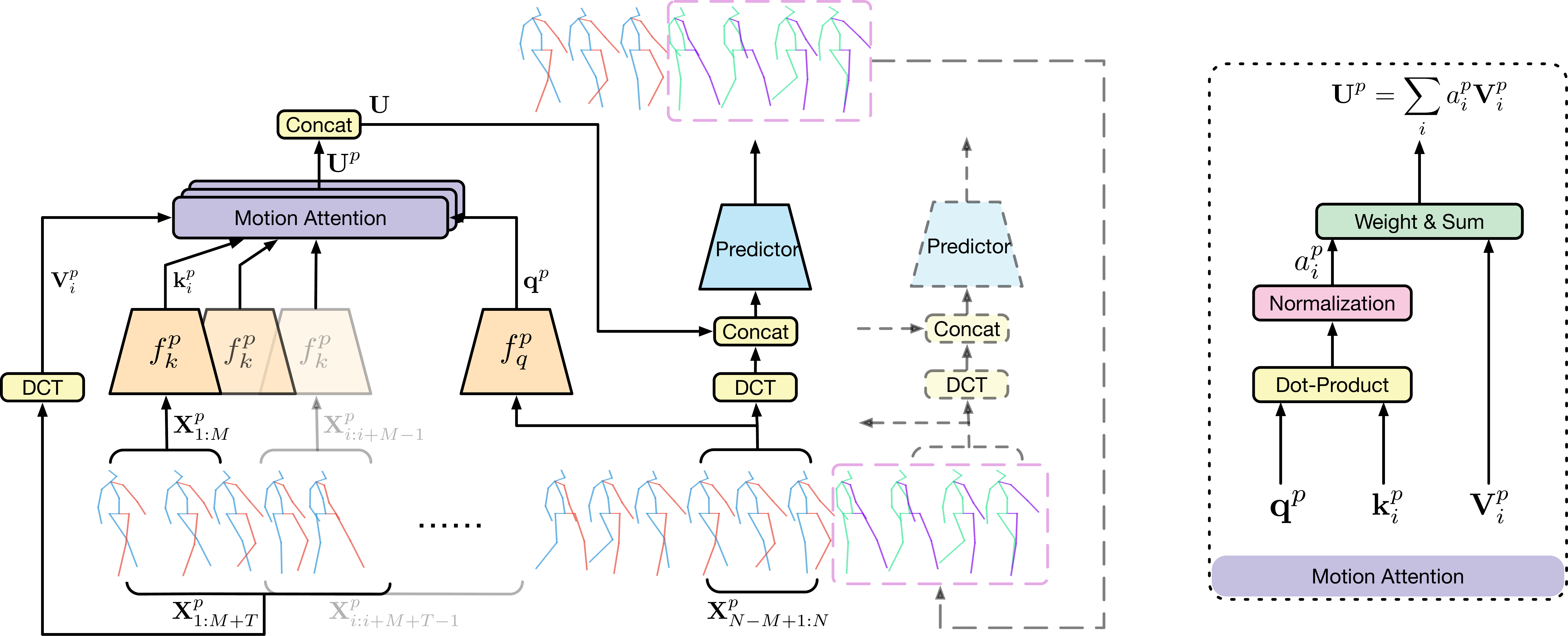}
      \caption{{\bf Overview of our motion attention pipeline.} The past poses are shown as blue and red skeletons and the predicted ones in green and purple. 
      The last observed $M$ poses are initially used as query. 
      For every $M$ consecutive poses in the history (key), we compute an attention score to weigh the DCT coefficients (values) of the corresponding sub-sequence. The weighted sum of such values is then concatenated with the DCT coefficients of the last observed sub-sequence to predict the future. At test time, to predict poses in the further future, we use the output of the predictor as input and predict future motion recursively (as illustrated by the dashed line).}
      \label{fig:net-structure}
\end{figure*}

\noindent{{\bf Feed-forward methods and long motion history encoding.}} 
In view of the drawbacks of RNNs, several works considered feed-forward networks as an alternative solution~\citep{Butepage_2017_CVPR,LiZLL18,mao2019learning}. In particular, \cite{Butepage_2017_CVPR} introduced a fully-connected network to process the recent pose history, investigating different strategies to encode temporal historical information via convolutions and exploiting the kinematic tree to encode spatial information. However, similarly to~\citep{JainZSS16}, and as discussed in~\citep{LiZLL18}, the use of a fixed tree structure does not reflect the motion synchronization across different, potentially distant, human body parts. To capture such dependencies, \cite{LiZLL18} built a convolutional sequence-to-sequence model processing a two-dimensional pose matrix whose columns represent the pose at every time step. This model was then used to extract a prior from long-term motion history, which, in conjunction with the more recent motion history, was used as input to an autoregressive network for future pose prediction. While more effective than the RNN-based frameworks, the manually-selected size of the convolutional window highly influences the temporal encoding.

Our work builds on our previous work~\citep{mao2019learning}, which showed that encoding the short-term history in frequency space using the DCT, followed by a GCN to encode spatial and temporal connections led to state-of-the-art performance for human motion prediction up to 1s. However, encoding long-term history in DCT yields an overly-general motion representation, leading to worse performance than using short-term history. In this paper, we overcome this drawback by introducing a~\emph{motion attention} based approach to human motion prediction. This allows us to capture the motion recurrence in the long-term history. Furthermore, in contrast to~\citep{LiZLL18}, whose encoding of past motions depends on the manually-defined size of the temporal convolution filters, our model dynamically adapts its history-based representation to the context of the current prediction.

\noindent{{\bf Attention models for human motion prediction.}} While attention-based neural networks are commonly employed for machine translation~\citep{vaswani2017attention,bahdanau2014neural}, their use for human motion prediction remains largely unexplored. 
The work of \cite{Tang_2018} constitutes an exception, incorporating an attention module to summarize the recent pose history, followed by an RNN-based prediction network. This work, however, uses frame-wise pose-based attention, which may lead to ambiguous motion, because static poses do not provide information about the motion direction and similar poses occur in significantly different motions. To overcome this, we propose to leverage~\emph{motion attention}. As evidenced by our experiments, this, combined with a feed-forward prediction network, allows us to outperform the state-of-the-art motion prediction frameworks.

In a similar spirit to our approach, the concurrent work of \cite{cai2020learning} leverages an attention-based transformer for human motion prediction. Nevertheless, their attention module mainly serves to model the global \emph{spatial} dependencies among the joint trajectories. By contrast, our motion attention aims to capture the motion repetitiveness in history, thus modeling \emph{temporal} motion dependencies. In addition to the attention-based module, \cite{cai2020learning} proposed to progressively predict the joint trajectories with a dictionary which stores the global motion patterns of training data. These two components, however, are orthogonal to the attention-based module. Our experiments demonstrate that our method outperforms that of~\cite{cai2020learning} with only the attention-based module and is comparable to the full-model of~\cite{cai2020learning}.

\section{Our Approach}
Let us now introduce our approach to human motion prediction. Let ${\bf X}_{1:N} = [{\bf x}_1,{\bf x}_2,{\bf x}_3,\cdots,{\bf x}_N]$ encode the motion history, consisting of $N$ consecutive human poses, where ${\bf x}_i\in \mathbf{R}^K$, with $K$ the number of parameters describing each pose, in our case 3D coordinates or angles of human joints.  Our goal is to predict the poses ${\bf X}_{N+1:N+T}$ for the future $T$ time steps. To this end, we introduce a motion attention model that allows us to form a future motion estimate by aggregating the long-term temporal information from the history.
We then combine this estimate with the latest observed motion and input this combination to a GCN-based feed-forward network that lets us learn the spatial and temporal dependencies in the data. 
Below, we discuss these two steps in detail.
\begin{figure*}[!ht]
    \centering
    \includegraphics[width=\linewidth]{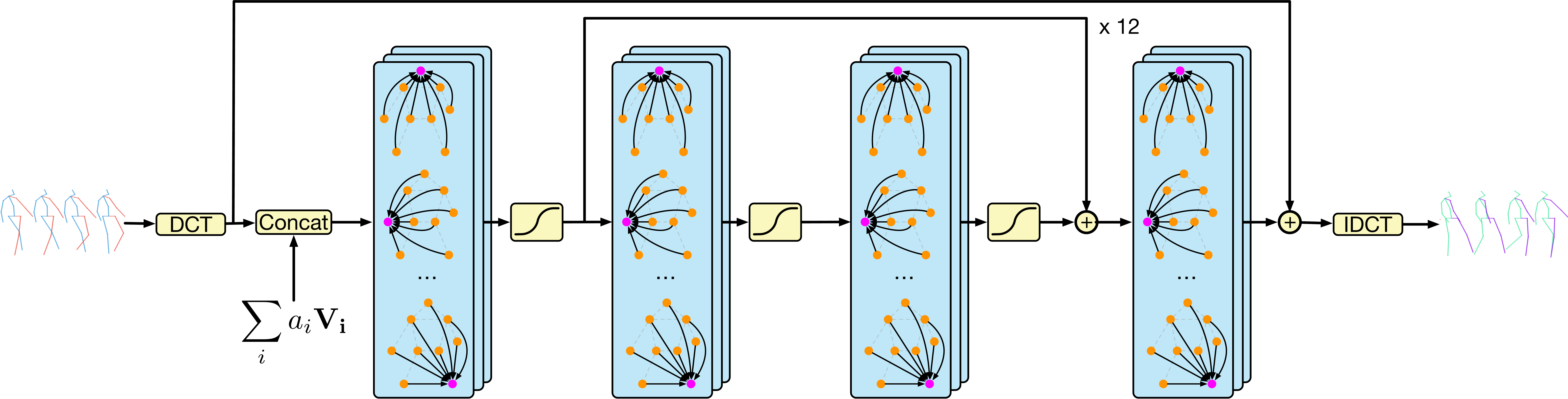}
    \caption{{\bf Predictor.} We first apply the DCT to encode temporal pose information in trajectory space. The DCT coefficients concatenated with the output of motion attention model are treated as features input to graph convolutional layers. In each layer, we depict how our framework aggregates information from multiple nodes via learned adjacency matrices.}
    \label{fig:pred_model}
\end{figure*}
\begin{figure*}[!ht]
    \centering
    \begin{tabular}{ccc}
    \includegraphics[width=0.3\linewidth]{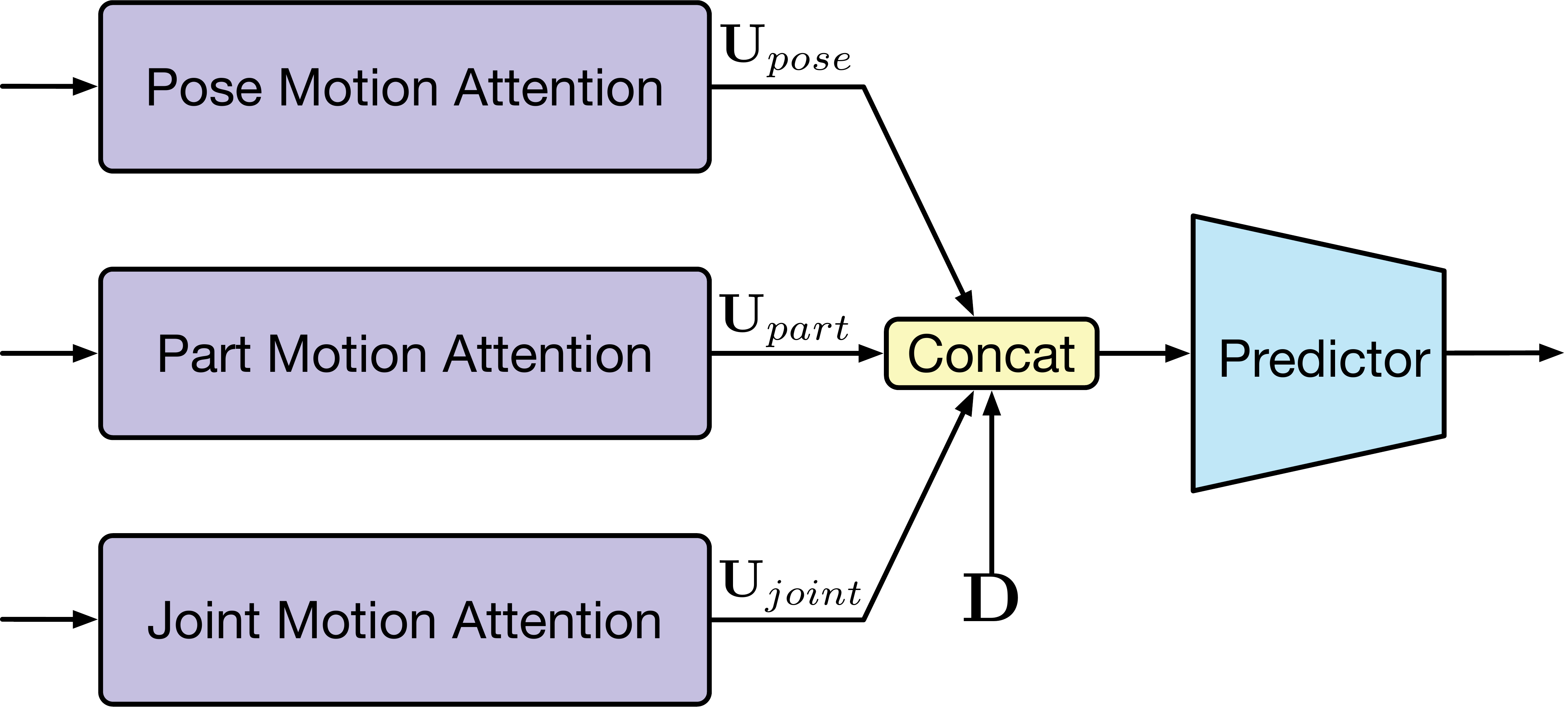} & 
    \includegraphics[width=0.33\linewidth]{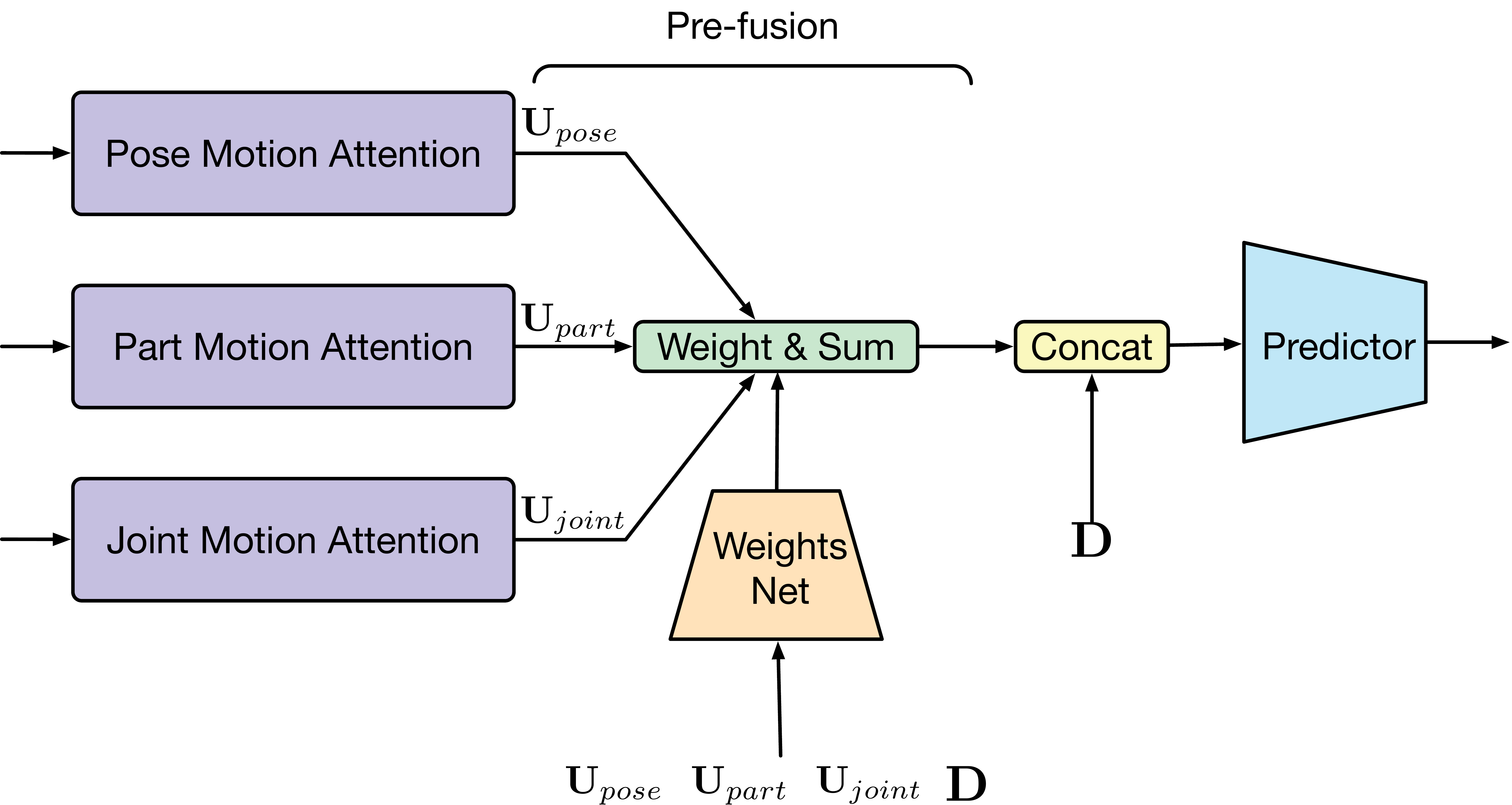} & 
    \includegraphics[width=0.33\linewidth]{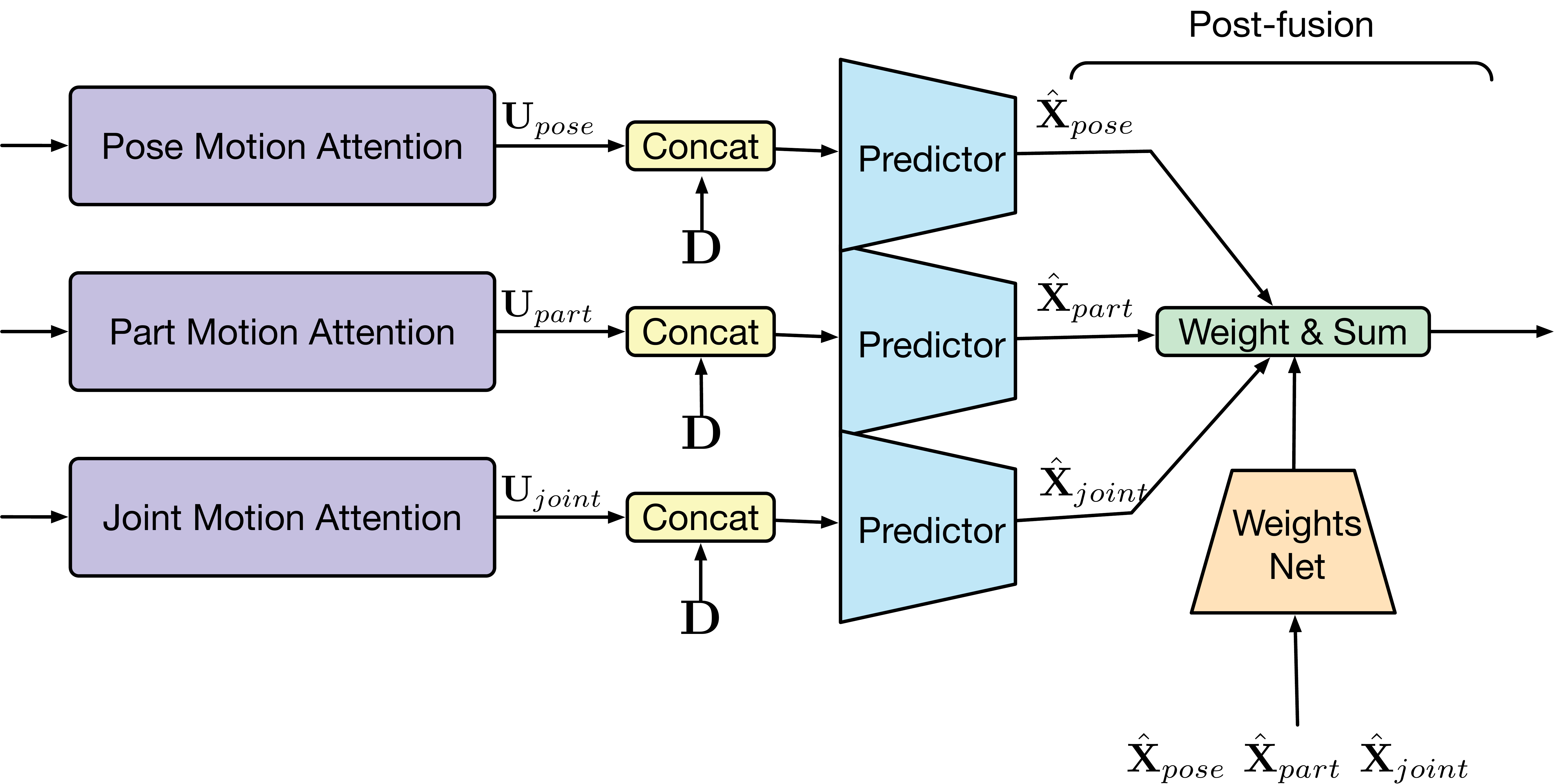}\\
    (a)&(b)&(c)
    \end{tabular}
    \caption{{\bf Different fusion model.} (a) Simply concatenate the outputs of all three motion models with the DCT coefficient ${\bf D}$. (b) Pre-fusion: The outputs of three motion models are first combined by a fusion model and fed into the predictor. (c) Post-fusion: The fusion process occurs after the predictions are made.}
    \label{fig:sele_model}
\end{figure*}
\subsection{Motion Attention Model}\label{sec:att-model}
As humans tend to repeat their motion across long time periods, our goal is to discover sub-sequences in the motion history that are similar to the current sub-sequence. We propose to achieve this via an attention model. 
To capture motion repetitiveness at different levels, we introduce a general framework that models attention on body parts. Specifically, a part can be the entire body, a human limb, e.g., the right arm, or an individual joint. This framework allows us to study different levels of attention, such as \emph{pose motion attention}, \emph{part motion attention} and \emph{joint motion attention}.

To this end, we first divide each human pose ${\bf x}_i\in\mathbf{R}^{K}$ into $P$ parts as
    $${\bf x}_i=\begin{bmatrix}
    {\bf x}_i^1\\
    {\bf x}_i^2\\ 
    {\bf x}_i^3\\
    \vdots\\
    {\bf x}_i^P
    \end{bmatrix}$$
where ${\bf x}_i^p\in\mathbf{R}^{K_p}$ concatenates the 3D coordinates (or rotation angles) of one body part and $\sum_{p=1}^{P}{K_p} = K$.
In particular, $P=1$ corresponds to treating the entire human pose as a single part, $P=N_J$, with $N_J$ is the number of skeleton joints, means that each joint acts as a part, whereas $P\in (1, N_J)$ ranges between these two extreme cases, grouping multiple joints into a part.

Following $\;$the $\;$machine $\;$translation $\;$formalism of \cite{vaswani2017attention}, we describe our attention model as a mapping from a~\emph{query} and a set of~\emph{key-value} pairs to an output. The output is a weighted sum of~\emph{values}, where the weight, or \emph{attention}, assigned to each value is a function of its corresponding~\emph{key} and of the~\emph{query}.~Mapping to our motion attention model, the~\emph{query} corresponds to a learned representation of the last observed sub-sequence, and the~\emph{key-value} pairs are treated as a dictionary within which~\emph{keys} are learned representations for historical sub-sequences and~\emph{values} are the corresponding learned future motion representations. Our motion attention model output is defined as the aggregation of these future motion representations based on~\emph{partial motion similarity} between the latest motion sub-sequence and historical sub-sequences.

In our context, we aim to compute attention from short sequences. To this end, we first divide the motion history of each body part ${\bf X}_{1:N}^{p} = [{\bf x}_1^{p},{\bf x}_2^{p},{\bf x}_3^{p},\cdots,{\bf x}_N^{p}]$, with $p\in\{1,2,\cdots, P\}$, into $N-M-T+1$ sub-sequences $\{{\bf X}_{i:i+M+T-1}^{p}\}_{i=1}^{N-M-T+1}$, each of which consists of $M+T$ consecutive body part poses. By using sub-sequences of length $M+T$, we assume that the motion predictor, which we will introduce in Section~\ref{sec:pred}, exploits the past $M$ frames to predict the future $T$ frames. 
We then take the first $M$ poses of each sub-sequence ${\bf X}_{i:i+M-1}^{p}$ to be a~\emph{key}, and the whole sub-sequence ${\bf X}_{i:i+M+T-1}^{p}$ is the corresponding~\emph{value}. Furthermore, we define the~\emph{query} as the latest sub-sequence ${\bf X}_{N-M+1:N}^{p}$ with length $M$.

To make the output of our attention model consistent with that of the final predictor, we map the resulting values to trajectory space using the DCT on the temporal dimension. That is, we take our final values to be the DCT coefficients $\{{\bf V}_i^{p}\}_{i=1}^{N-M-T+1}$, where ${\bf V}_i^{p}\in \mathbf{R}^{K_{p} \times (M+T)}$. Each row of ${\bf V}_i^{p}$ contains the DCT coefficients of one joint coordinate sequence. In practice, we can truncate some high frequencies to avoid predicting jittery motion.

As depicted by Fig.~\ref{fig:net-structure}, the query and keys are used to compute attention scores, which then act as weights to combine the corresponding values. To this end, we first map the query and keys to vectors of the same dimension $d$ by two functions $f_{q}^{p}:\mathbf{R}^{K_{p}\times M}\rightarrow \mathbf{R}^{d}$ and $f_{k}^{p}:\mathbf{R}^{K_{p}\times M}\rightarrow \mathbf{R}^{d}$ modeled with neural networks. This can be expressed as
\begin{equation}
    {\bf q}^{p} = f_{q}^{p}({\bf X}_{N-M+1:N}^{p})\;,
\end{equation}
\begin{equation}
     {\bf k}_{i}^{p} = f_{k}^{p}({\bf X}_{i:i+M-1}^{p})\;,
\end{equation}
where ${\bf q}^{p},{\bf k}_i^{p}\in \mathbf{R}^{d}$, $i \in \{1,2,\cdots,N-M-T+1\}$, and $p \in \{1,2,\cdots,P\}$.
For each key, we then compute an attention score as
\begin{equation}
a_i^{p} = \frac{{\bf q}^{p}{{\bf k}_{i}^{p}}^T}{\sum_{i=1}^{N-M-T+1}{{\bf q}^{p}{{\bf k}_{i}^{p}}^T}}\;.
\end{equation}
Note that, instead of the softmax function which is commonly used in attention mechanisms, we simply normalize the attention scores by their sum, which we found to avoid the gradient vanishing problem that may occur when using a softmax. While this division only enforces the sum of the attention scores to be $1$, we further restrict the outputs of $f_q^{p}$ and $f_k^{p}$ to be non-negative with ReLU to avoid obtaining negative attention scores.

We then compute the output of the attention model for each body part as the weighed sum of values, i.e.,
\begin{equation}
\centering
    {\bf U}_{p} = \sum_{i=1}^{N-M-T+1}{a_i^{p}{\bf V}_i^{p}}\;,
\end{equation}
where ${\bf U}_{p}\in \mathbf{R}^{K_{p} \times (M+T)}$. The final output for the whole body is the concatenation of those for all body parts ${\bf U}=[{\bf U}_{1}^T,{\bf U}_{2}^T,\cdots,{\bf U}_{P}^T]^T$ with ${\bf U}\in \mathbf{R}^{K\times(M+T)}$. This initial estimate is then combined with the latest sub-sequence and processed by the prediction model described below to generate future poses $\hat{{\bf X}}_{N+1:N+T}$. 

At test time, to generate longer future motion, we augment the motion history with the last predictions and update the query with the latest sub-sequence in the augmented motion history, and the key-value pairs accordingly. These updated entities are then used for the next prediction step.

\subsection{Prediction Model}\label{sec:pred}
To predict the future motion, we reuse the motion prediction model we introduced in~\citep{mao2019learning} as shown in Fig.~\ref{fig:pred_model}. Specifically, as mentioned above, we use a DCT-based representation to encode the temporal information for each joint coordinate or angle and GCNs with learnable adjacency matrices to capture the spatial dependencies among the coordinates or angles.

\begin{table*}[ht]
\centering
\resizebox{0.95\textwidth}{!}{%
\begin{tabular}{ccccc|cccc|cccc|cccc}
& \multicolumn{4}{c}{Walking} & \multicolumn{4}{c}{Eating} & \multicolumn{4}{c}{Smoking} & \multicolumn{4}{c}{Discussion} \\
milliseconds & 80   & 160  & 320  & 400  & 80   & 160  & 320  & 400  & 80   & 160  & 320  & 400  & 80   & 160  & 320  & 400\\\hline
Res. Sup.~\citep{Martinez_2017_CVPR} &  23.2 & 40.9 & 61.0 & 66.1 & 16.8 & 31.5 & 53.5 & 61.7 & 18.9 & 34.7 & 57.5 & 65.4 & 25.7 & 47.8 & 80.0 & 91.3 \\
convSeq2Seq~\citep{LiZLL18} & 17.7 & 33.5 & 56.3 & 63.6 & 11.0 & 22.4 & 40.7 & 48.4 & 11.6 & 22.8 & 41.3 & 48.9 & 17.1 & 34.5 & 64.8 & 77.6  \\
LTD-50-25\citep{mao2019learning} & 12.3 & 23.2 & 39.4 & 44.4 & 7.8 & 16.3 & 31.3 & 38.6 & 8.2 & 16.8 & 32.8 & 39.5 & 11.9 & 25.9 & 55.1 & 68.1 \\
LTD-10-25\citep{mao2019learning} & 12.6 & 23.6 & 39.4 & 44.5 & 7.7 & 15.8 & 30.5 & 37.6 & 8.4 & 16.8 & 32.5 & 39.5 & 12.2 & 25.8 & 53.9 & 66.7 \\
LTD-10-10\citep{mao2019learning} & 11.1 & 21.4 & 37.3 & 42.9 & 7.0 & 14.8 & 29.8 & 37.3 & 7.5 & 15.5 & 30.7 & 37.5 & 10.8 & 24.0 & 52.7 & 65.8 \\\hline
Pose Motion Att.~\citep{mao2020history} & 10.0 & 19.5 & 34.2 & 39.8 & 6.4 & 14.0 & 28.7 & 36.2 & 7.0 & 14.9 & 29.9 & 36.4 & 10.2 & 23.4 & 52.1 & 65.4\\
Motion Att. + Post-fusion & \textbf{9.9} & \textbf{19.3} & \textbf{33.7} & \textbf{39.0} & \textbf{6.2} & \textbf{13.7} & \textbf{28.1} & \textbf{35.3} & \textbf{6.8} & \textbf{14.5} & \textbf{29.0} & \textbf{35.5} & \textbf{9.9} & \textbf{22.8} & \textbf{51.0} & \textbf{64.0}\\\hline\hline
LPJ (8 Sub-seq)~\citep{cai2020learning} & \textbf{7.9} & 14.5 & 29.1 & 34.5 & 8.4 & 18.1 & \textbf{37.4} & 45.3 & \textbf{6.8} & \textbf{13.2} & \textbf{24.1} & \textbf{27.5} & \textbf{8.3} & 21.7 & 43.9 & 48.0\\\hline
Motion Att. + Post-fusion (8 Sub-seq) & \textbf{7.9} & \textbf{14.4} & \textbf{25.5} & \textbf{29.6} & \textbf{7.9} & \textbf{17.5} & \textbf{37.4} & \textbf{45.2} & 7.0 & 14.3 & 25.4 & 29.0 & 8.6 & \textbf{20.4} & \textbf{37.7} & \textbf{43.6}\\\hline\hline
& \multicolumn{4}{c}{Directions} & \multicolumn{4}{c}{Greeting} & \multicolumn{4}{c}{Phoning} & \multicolumn{4}{c}{Posing}\\
milliseconds& 80 & 160 & 320 & 400 & 80 & 160 & 320 & 400 & 80 & 160 & 320 & 400 & 80 & 160 & 320 & 400\\\hline
Res. Sup.~\citep{Martinez_2017_CVPR} & 21.6 & 41.3 & 72.1 & 84.1 & 31.2 & 58.4 & 96.3 & 108.8 & 21.1 & 38.9 & 66.0 & 76.4 & 29.3 & 56.1 & 98.3 & 114.3 \\
convSeq2Seq~\citep{LiZLL18} & 13.5 & 29.0 & 57.6 & 69.7 & 22.0 & 45.0 & 82.0 & 96.0 & 13.5 & 26.6 & 49.9 & 59.9 & 16.9 & 36.7 & 75.7 & 92.9\\
LTD-50-25\citep{mao2019learning} & 8.8 & 20.3 & 46.5 & 58.0 & 16.2 & 34.2 & 68.7 & 82.6 & 9.8 & 19.9 & 40.8 & 50.8 & 12.2 & 27.5 & 63.1 & 79.9\\
LTD-10-25\citep{mao2019learning} & 9.2 & 20.6 & 46.9 & 58.8 & 16.7 & 33.9 & 67.5 & 81.6 & 10.2 & 20.2 & 40.9 & 50.9 & 12.5 & 27.5 & 62.5 & 79.6\\
LTD-10-10\citep{mao2019learning} & 8.0 & 18.8 & 43.7 & \textbf{54.9} & 14.8 & 31.4 & 65.3 & 79.7 & 9.3 & 19.1 & 39.8 & 49.7 & 10.9 & 25.1 & 59.1 & 75.9 \\\hline
Pose Motion Att.~\citep{mao2020history} & 7.4 & 18.4 & 44.5 & 56.5 & 13.7 & 30.1 & 63.8 & 78.1 & 8.6 & 18.3 & 39.0 & 49.2 & 10.2 & 24.2 & 58.5 & 75.8 \\
Motion Att. + Post-fusion & \textbf{7.2} & \textbf{18.0} & \textbf{43.4} & 55.0 & \textbf{13.6} & \textbf{29.9} & \textbf{62.9} & \textbf{77.2} & \textbf{8.4} & \textbf{18.0} & \textbf{38.3} & \textbf{48.4} & \textbf{9.8} & \textbf{23.7} & \textbf{57.8} & \textbf{74.9}\\\hline\hline
LPJ (8 Sub-seq)~\citep{cai2020learning} & \textbf{11.1} & \textbf{22.7} & \textbf{48.0} & \textbf{58.4} & 13.2 & 28.0 & \textbf{64.5} & \textbf{77.9} & \textbf{10.8} & \textbf{19.6} & \textbf{37.6} & 46.8 & 8.3 & 22.8 & 65.6 & 81.8 \\\hline
Motion Att. + Post-fusion (8 Sub-seq) & 11.3 & 22.9 & 50.6 & 62.6 & \textbf{12.9} & \textbf{26.6} & 68.2 & 85.4 & 11.2 & \textbf{19.6} & 37.7 & \textbf{44.1} & \textbf{7.7} & \textbf{21.3} & \textbf{62.2} & \textbf{78.7} \\\hline
&\multicolumn{4}{c}{Purchases}&\multicolumn{4}{c}{Sitting}& \multicolumn{4}{c}{Sitting Down} & \multicolumn{4}{c}{Taking Photo} \\
milliseconds& 80 & 160 & 320 & 400 & 80 & 160 & 320 & 400 & 80 & 160 & 320 & 400 & 80 & 160 & 320 & 400\\\hline
Res. Sup.~\citep{Martinez_2017_CVPR} & 28.7 & 52.4 & 86.9 & 100.7 & 23.8 & 44.7 & 78.0 & 91.2 & 31.7 & 58.3 & 96.7 & 112.0 & 21.9 & 41.4 & 74.0 & 87.6\\
convSeq2Seq~\citep{LiZLL18} & 20.3 & 41.8 & 76.5 & 89.9 & 13.5 & 27.0 & 52.0 & 63.1 & 20.7 & 40.6 & 70.4 & 82.7 & 12.7 & 26.0 & 52.1 & 63.6\\
LTD-50-25\citep{mao2019learning} & 15.2 & 32.9 & 64.9 & 78.1 & 10.4 & 21.9 & 46.6 & 58.3 & 17.1 & 34.2 & 63.6 & 76.4 & 9.6 & 20.3 & 43.3 & 54.3\\
LTD-10-25\citep{mao2019learning} &  15.5 & 32.3 & 63.6 & 77.3 & 10.4 & 21.4 & 45.4 & 57.3 & 17.0 & 33.4 & 61.6 & 74.4 & 9.9 & 20.5 & 43.8 & 55.2\\
LTD-10-10\citep{mao2019learning} & 13.9 & 30.3 & 62.2 & 75.9 & 9.8 & 20.5 & 44.2 & 55.9 & 15.6 & 31.4 & 59.1 & 71.7 & 8.9 & 18.9 & 41.0 & 51.7\\\hline
Pose Motion Att.~\citep{mao2020history} & 13.0 & 29.2 & 60.4 & 73.9 & 9.3 & 20.1 & 44.3 & 56.0 & 14.9 & 30.7 & 59.1 & 72.0 & 8.3 & 18.4 & 40.7 & 51.5\\
Motion Att. + Post-fusion & \textbf{12.8} & \textbf{28.7} & \textbf{59.4} & \textbf{72.8} & \textbf{9.1} & \textbf{19.7} & \textbf{43.7} & \textbf{55.4} & \textbf{14.7} & \textbf{30.4} & \textbf{58.4} & \textbf{71.3} & \textbf{8.2} & \textbf{18.1} & \textbf{40.2} & \textbf{51.1}\\\hline\hline
LPJ (8 Sub-seq)~\citep{cai2020learning} & 18.5 & 38.1 & 61.8 & 69.6 & \textbf{9.5} & \textbf{23.9} & \textbf{49.8} & \textbf{61.8} & 11.2 & 29.9 & 59.8 & 68.4 & 6.3 & \textbf{14.5} & 38.8 & \textbf{49.4} \\\hline
Motion Att. + Post-fusion (8 Sub-seq) & \textbf{18.1} & \textbf{36.8} & \textbf{58.4} & \textbf{67.9} & 9.9 & 24.3 & 53.8 & 66.3 & \textbf{10.4} & \textbf{26.6} & \textbf{54.6} & \textbf{66.3} & \textbf{5.9} & 14.8 & \textbf{38.0} & \textbf{49.4}\\\hline
& \multicolumn{4}{c}{Waiting} & \multicolumn{4}{c}{Walking Dog}&\multicolumn{4}{c}{Walking Together}&\multicolumn{4}{c}{Average} \\
milliseconds& 80 & 160 & 320 & 400 & 80 & 160 & 320 & 400 & 80 & 160 & 320 & 400 & 80 & 160 & 320 & 400\\\hline
Res. Sup.~\citep{Martinez_2017_CVPR}  & 23.8 & 44.2 & 75.8 & 87.7 & 36.4 & 64.8 & 99.1 & 110.6 & 20.4 & 37.1 & 59.4 & 67.3 & 25.0 & 46.2 & 77.0 & 88.3 \\
convSeq2Seq~\citep{LiZLL18} & 14.6 & 29.7 & 58.1 & 69.7 & 27.7 & 53.6 & 90.7 & 103.3 & 15.3 & 30.4 & 53.1 & 61.2 & 16.6 & 33.3 & 61.4 & 72.7  \\
LTD-50-25\citep{mao2019learning} & 10.4 & 22.1 & 47.9 & 59.2 & 22.8 & 44.7 & 77.2 & 88.7 & 10.3 & 21.2 & 39.4 & 46.3 & 12.2 & 25.4 & 50.7 & 61.5\\
LTD-10-25\citep{mao2019learning} & 10.5 & 21.6 & 45.9 & 57.1 & 22.9 & 43.5 & 74.5 & 86.4 & 10.8 & 21.7 & 39.6 & 47.0 & 12.4 & 25.2 & 49.9 & 60.9\\
LTD-10-10\citep{mao2019learning} & 9.2 & 19.5 & 43.3 & 54.4 & 20.9 & 40.7 & 73.6 & 86.6 & 9.6 & 19.4 & 36.5 & 44.0 & 11.2 & 23.4 & 47.9 & 58.9\\\hline
Pose Motion Att.~\citep{mao2020history} & 8.7 & 19.2 & 43.4 & 54.9 & 20.1 & 40.3 & 73.3 & 86.3 & 8.9 & 18.4 & 35.1 & 41.9 & 10.4 & 22.6 & 47.1 & 58.3 \\
Motion Att. + Post-fusion & \textbf{8.4} & \textbf{18.7} & \textbf{42.5} & \textbf{53.8} & \textbf{19.6} & \textbf{39.5} & \textbf{71.7} & \textbf{84.1} & \textbf{8.5} & \textbf{17.9} & \textbf{34.3} & \textbf{41.1} & \textbf{10.2} & \textbf{22.2} & \textbf{46.3} & \textbf{57.3}\\\hline\hline
LPJ (8 Sub-seq)~\citep{cai2020learning} & \textbf{8.4} & \textbf{21.5} & \textbf{53.9} & \textbf{69.8} & \textbf{22.9} & \textbf{50.4} & 100.8 & 119.8 & 8.7 & 18.3 & 34.2 & 44.1 & \textbf{10.7} & 23.8 & 50.0 & 60.2\\\hline
Motion Att. + Post-fusion (8 Sub-seq) & 9.0 & 22.5 & 55.7 & 71.1 & 29.5 & 54.8 & \textbf{100.3} & \textbf{119.0} & \textbf{8.0} & \textbf{17.6} & \textbf{33.2} & \textbf{42.0} & 11.0 & \textbf{23.6} & \textbf{49.2} & \textbf{60.0}\\\hline
\end{tabular}
}
\caption{Short-term prediction of 3D joint positions on H3.6M. The error is measured in millimeter. 
For ``LTD'', we use the number of observed frames and that of future frames to predict during training to distinguish different models. For instance, ``LTD-50-25'' means the model is trained to observe past 50 frames and predict future 25 frames. 
Following QuaterNet~\citep{pavllo2019modeling}, we report the average error on 256 sub-sequences except for those with ``(8 Sub-seq)" after their method names is averaging over 8 sub-sequences per action. Our approach achieves state of the art performance across all 15 actions at almost all time horizons, especially for actions with a clear repeated history, such as ``Walking". The proposed extension ``Post-fusion" further improves the results compared to the base model ``Pose Motion Att.".}
\label{tab:h36_short_3d}
\end{table*}

\begin{table*}[ht]
\centering
\resizebox{\textwidth}{!}{%
\begin{tabular}{ccccc|cccc|cccc|cccc|cccc}
& \multicolumn{4}{c}{Walking} & \multicolumn{4}{c}{Eating} & \multicolumn{4}{c}{Smoking} & \multicolumn{4}{c}{Discussion} & \multicolumn{4}{c}{Average}\\
milliseconds & 80   & 160  & 320  & 400  & 80   & 160  & 320  & 400  & 80   & 160  & 320  & 400  & 80   & 160  & 320  & 400   & 80   & 160  & 320  & 400\\\hline
LPJ w/ att., w/o prog., w/o dict. ~\citep{cai2020learning} & 10.5 & 17.1 & 31.9 & 35.7 & 10.1 & 21.2 & 40.7 & 47.5 & 8.6 & 15.9 & 26.5 & 30.4 & 10.6 & 24.1 & 47.5 & 51.3 & 9.9 & 19.5 & 36.6 & 41.2 \\\hline
Pose Motion Att. & 8.1 & 14.7 & \textbf{25.0} & \textbf{29.4} & 8.2 & 18.2 & 38.6 & 46.9 & 7.0 & 14.5 & 25.9 & 29.2 & 8.8 & 21.7 & 40.0 & 45.9 & 8.1 & 17.3 & 32.4 & 37.9 \\
Motion Att. + Post-fusion & \textbf{7.9} & \textbf{14.4} & 25.5 & 29.6 & \textbf{7.9} & \textbf{17.5} & \textbf{37.4} & \textbf{45.2} & \textbf{7.0} & \textbf{14.3} & \textbf{25.4} & \textbf{29.0} & \textbf{8.6} & \textbf{20.4} & \textbf{37.7} & \textbf{43.6} & \textbf{7.9} & \textbf{16.7} & \textbf{31.5} & \textbf{36.9}\\\hline
\end{tabular}
}
\caption{Short-term prediction of 3D joint positions on 4 actions of H3.6. Since LPJ~\citep{cai2020learning} only provide their results on 4 action of H3.6M, we compare our results with that of LPJ~\citep{cai2020learning} on these actions. The ``att.", ``prog." and ``dict." refer to attention-based prediction, progressive prediction and motion dictionary which are the 3 components proposed in LPJ~\citep{cai2020learning}. With only the attention module, our method outperforms that of LPJ~\citep{cai2020learning} by a large margin on all cases.}
\label{tab:h36_short_3d_no_prog}
\end{table*}
\noindent{{\bf Temporal encoding}.} Given a motion sequence ${\bf X}_{1:L}$, whose $k^{th}$ row can be expressed as $[x_{k,1},x_{k,2},\cdots,x_{k,L}]$, the corresponding $l^{th}$ DCT coefficient of this row is computed as
\begin{equation}
\resizebox{0.92\linewidth}{!}{%
$C_{k,l} = \sqrt{\frac{2}{L}}\sum_{n=1}^{L}x_{k,n}\frac{1}{\sqrt{1+\delta_{l1}}}\cos\left(\frac{\pi}{2L}(2n-1)(l-1)\right)\;,$
}\label{eq:dct}
\end{equation}
where $l\in\{1,2,\cdots,L\}$ and $\delta_{ij}$ denotes the Kronecker delta function, i.e., 
\begin{equation}
  \delta_{ij} = \begin{cases}
  1 & \text{if}\ i=j\\
  0 & \text{if}\ i\neq j.
  \end{cases}
\end{equation}
Given such coefficients, the original pose representation (coordinates or angles) can be obtained via the Inverse Discrete Cosine Transform (IDCT) as
\begin{equation}
\resizebox{0.92\linewidth}{!}{%
    $x_{k,n} =\sqrt{\frac{2}{L}}\sum_{l=1}^{L}C_{k,l}\frac{1}{\sqrt{1+\delta_{l1}}}\cos\left(\frac{\pi}{2L}(2n-1)(l-1)\right)\;,$
    }
\end{equation}
where $n\in\{1,2,\cdots,L\}$.

To predict future poses ${\bf X}_{N+1:N+T}$, we make use of the latest sub-sequence ${\bf X}_{N-M+1:N}$, which is also the~\emph{query} in the attention model. Adopting the same padding strategy as in~\citep{mao2019learning},  we replicate the last observed pose ${\bf X}_N$ $T$ times to generate a sequence of length $M+T$ and the DCT coefficients of this sequence are denoted as ${\bf D}\in \mathbf{R}^{K\times (M+T)}$. We then aim to predict DCT coefficients of the future sequence ${\bf X}_{N-M+1:N+T}$ given ${\bf D}$ and the attention model's output ${\bf U}$.

\noindent{{\bf Spatial encoding}.} To capture spatial dependencies between different joint coordinates or angles, we regard the human body as a fully-connected graph with $K$ nodes. 
The input to a graph convolutional layer $m$ is a matrix ${\bf H}^{(m)}\in \mathbf{R}^{K\times F}$, where each row is the $F$ dimensional feature vector of one node. For example, for the first layer, the network takes as input the $K \times 2(M+T)$ matrix that concatenates ${\bf D}$ and ${\bf U}$. A graph convolutional layer then outputs a $K\times\hat{F}$ matrix of the form
\begin{equation}
    {\bf H}^{(m+1)} = \sigma({\bf A}^{(m)}{\bf H}^{(m)}{\bf W}^{(m)})\;,
\end{equation}
where ${\bf A}^{(m)}\in \mathbf{R}^{K\times K}$ is the trainable adjacency matrix of layer $m$, representing the strength of the connectivity between nodes, ${\bf W}^{(m)}\in \mathbf{R}^{F \times \hat{F}}$ also encodes trainable weights but used to extract features, and $\sigma(\cdot)$ is an activation function, such as $tanh(\cdot)$. We stack several such layers to form our GCN-based predictor.

Given ${\bf D}$ and ${\bf U}$, the predictor learns a residual between the DCT coefficients ${\bf D}$ of the padded sequence and those of the true sequence. By applying IDCT to the predicted DCT coefficients, we obtain the coordinates or angles $\hat{{\bf X}}_{N-M+1:N+T}$, whose last $T$ poses $\hat{{\bf X}}_{N+1:N+T}$ are predictions in the future.

\subsection{Fusion Model}\label{sec:selec_model} 
As mentioned before, different activities/sequences may benefit from using attention at different levels, i.e., full body, parts, or individual joints. To model this, we introduce a fusion model that automatically combines different attention model and obtains the best-suited attention level for the current context.
Specifically, partitioning the human skeleton into full pose, body parts and individual joints, corresponding to different choices of $P$ in Section~\ref{sec:att-model}, we compute the motion attentions ${\bf U}_{pose}$, ${\bf U}_{part}$, and ${\bf U}_{joint}$, respectively, and treat them as motion priors.
We then study the three different ways to exploit these motion priors depicted by Fig.~\ref{fig:sele_model}. The first one (Fig.~\ref{fig:sele_model}(a)) consists of simply concatenating them with the DCT coefficients ${\bf D}$ of the padded sequence before being fed to the predictor. The other two ways both involve training a fusion model which outputs 3 normalized weights, one for each type of motion prior. The difference is where it is applied. For \emph{pre-fusion} shown in Fig.~\ref{fig:sele_model}(b), the fusion model is used to fuse the outputs of the motion models before fed into the predictor, while in \emph{post-fusion}, the fusion model is trained to combine the predictions from three different predictors given different level of motion attention outputs.
As verified by our experiments, the post-fusion model of Fig.~\ref{fig:sele_model}(c) yields the best performance, and we therefore adopt it for our approach.

\begin{table*}[ht]
\centering
\resizebox{0.9\textwidth}{!}{%
\begin{tabu}{ccccc|cccc|cccc|cccc}
  & \multicolumn{4}{c}{Walking} & \multicolumn{4}{c}{Eating} & \multicolumn{4}{c}{Smoking} & \multicolumn{4}{c}{Discussion} \\
 milliseconds & 560 & 720 & 880 & 1000 & 560 & 720 & 880 & 1000 & 560 & 720 & 880 & 1000 & 560 & 720 & 880 & 1000\\\hline
Res. Sup.~\citep{Martinez_2017_CVPR} & 71.6 & 72.5 & 76.0 & 79.1 & 74.9 & 85.9 & 93.8 & 98.0 & 78.1 & 88.6 & 96.6 & 102.1 & 109.5 & 122.0 & 128.6 & 131.8\\
convSeq2Seq~\citep{LiZLL18} & 72.2 & 77.2 & 80.9 & 82.3 & 61.3 & 72.8 & 81.8 & 87.1 & 60.0 & 69.4 & 77.2 & 81.7 & 98.1 & 112.9 & 123.0 & 129.3\\
LTD-50-25\citep{mao2019learning} & 50.7 & 54.4 & 57.4 & 60.3 & 51.5 & 62.6 & 71.3 & 75.8 & 50.5 & 59.3 & 67.1 & 72.1 & 88.9 & 103.9 & 113.6 & 118.5\\
LTD-10-25\citep{mao2019learning} & 51.8 & 56.2 & 58.9 & 60.9 & 50.0 & 61.1 & 69.6 & 74.1 & 51.3 & 60.8 & 68.7 & 73.6 & 87.6 & 103.2 & 113.1 & 118.6\\
LTD-10-10\citep{mao2019learning} & 53.1 & 59.9 & 66.2 & 70.7 & 51.1 & 62.5 & 72.9 & 78.6 & 49.4 & 59.2 & 66.9 & 71.8 & 88.1 & 104.5 & 115.5 & 121.6\\\hline
Pose Motion Att.~\citep{mao2020history} & 47.4 & 52.1 & 55.5 & 58.1 & 50.0 & 61.4 & 70.6 & 75.7 & 47.6 & 56.6 & 64.4 & 69.5 & 86.6 & 102.2 & 113.2 & 119.8\\ 
Motion Att. + Post-fusion & \textbf{46.2} & \textbf{51.0} & \textbf{54.4} & \textbf{57.1} & \textbf{48.6} & \textbf{59.9} & \textbf{68.9} & \textbf{73.7} & \textbf{46.5} & \textbf{55.5} & \textbf{63.4} & \textbf{68.7} & \textbf{85.2} & \textbf{100.9} & \textbf{111.6} & \textbf{117.5} \\\hline
& \multicolumn{4}{c}{Directions} & \multicolumn{4}{c}{Greeting} & \multicolumn{4}{c}{Phoning} & \multicolumn{4}{c}{Posing}\\
 milliseconds & 560 & 720 & 880 & 1000 & 560 & 720 & 880 & 1000 & 560 & 720 & 880 & 1000 & 560 & 720 & 880 & 1000\\\hline
Res. Sup.~\citep{Martinez_2017_CVPR} & 101.1 & 114.5 & 124.5 & 129.1 & 126.1 & 138.8 & 150.3 & 153.9 & 94.0 & 107.7 & 119.1 & 126.4 & 140.3 & 159.8 & 173.2 & 183.2 \\
convSeq2Seq~\citep{LiZLL18} & 86.6 & 99.8 & 109.9 & 115.8 & 116.9 & 130.7 & 142.7 & 147.3 & 77.1 & 92.1 & 105.5 & 114.0 & 122.5 & 148.8 & 171.8 & 187.4 \\
LTD-50-25\citep{mao2019learning} & 74.2 & 88.1 & 99.4 & \textbf{105.5} & 104.8 & 119.7 & 132.1 & 136.8 & 68.8 & 83.6 & 96.8 & 105.1 & 110.2 & 137.8 & 160.8 & 174.8 \\
LTD-10-25\citep{mao2019learning} & 76.1 & 91.0 & 102.8 & 108.8 & 104.3 & 120.9 & 134.6 & 140.2 & 68.7 & 84.0 & 97.2 & 105.1 & 109.9 & 136.8 & 158.3 & \textbf{171.7} \\
LTD-10-10\citep{mao2019learning} & \textbf{72.2} & \textbf{86.7} & \textbf{98.5} & 105.8 & 103.7 & 120.6 & 134.7 & 140.9 & 67.8 & 83.0 & 96.4 & 105.1 & 107.6 & 136.1 & 159.5 & 175.0 \\\hline
Pose Motion Att.~\citep{mao2020history} & 73.9 & 88.2 & 100.1 & 106.5 & 101.9 & 118.4 & 132.7 & 138.8 & 67.4 & 82.9 & 96.5 & 105.0 & 107.6 & 136.8 & 161.4 & 178.2 \\
Motion Att. + Post-fusion & 72.4 & 87.4 & 99.3 & 105.7 & \textbf{100.5} & \textbf{116.5} & \textbf{130.7} & \textbf{136.7} & \textbf{66.5} & \textbf{82.3} & \textbf{95.8} & \textbf{104.6} & \textbf{105.8} & \textbf{134.1} & \textbf{157.5} & 172.9\\\hline
&\multicolumn{4}{c}{Purchases}&\multicolumn{4}{c}{Sitting} & \multicolumn{4}{c}{Sitting Down} & \multicolumn{4}{c}{Taking Photo}\\
 milliseconds & 560 & 720 & 880 & 1000 & 560 & 720 & 880 & 1000 & 560 & 720 & 880 & 1000 & 560 & 720 & 880 & 1000\\\hline
Res. Sup.~\citep{Martinez_2017_CVPR} & 122.1 & 137.2 & 148.0 & 154.0 & 113.7 & 130.5 & 144.4 & 152.6 & 138.8 & 159.0 & 176.1 & 187.4 & 110.6 & 128.9 & 143.7 & 153.9\\
convSeq2Seq~\citep{LiZLL18} & 111.3 & 129.1 & 143.1 & 151.5 & 82.4 & 98.8 & 112.4 & 120.7 & 106.5 & 125.1 & 139.8 & 150.3 & 84.4 & 102.4 & 117.7 & 128.1\\
LTD-50-25\citep{mao2019learning} & 99.2 & 114.9 & 127.1 & 134.9 & 79.2 & 96.2 & 110.3 & 118.7 & 100.2 & 118.2 & 133.1 & 143.8 & 75.3 & 93.5 & 108.4 & 118.8\\
LTD-10-25\citep{mao2019learning} & 99.4 & 114.9 & 127.9 & 135.9 & 78.5 & 95.7 & 110.0 & 118.8 & 99.5 & 118.5 & 133.6 & 144.1 & 76.8 & 95.3 & 110.3 & 120.2\\
LTD-10-10\citep{mao2019learning} & 98.3 & 115.1 & 130.1 & 139.3 & 76.4 & 93.1 & 106.9 & 115.7 & 96.2 & 115.2 & 130.8 & 142.2 & 72.5 & 90.9 & 105.9 & 116.3\\\hline
Pose Motion Att.~\citep{mao2020history} & 95.6 & 110.9 & 125.0 & 134.2 & 76.4 & 93.1 & 107.0 & 115.9 & 97.0 & 116.1 & 132.1 & 143.6 & 72.1 & 90.4 & 105.5 & 115.9\\
Motion Att. + Post-fusion & \textbf{94.5} & \textbf{110.2} & \textbf{124.4} & \textbf{133.1} & \textbf{75.8} & \textbf{92.3} & \textbf{106.0} & \textbf{115.0} & \textbf{96.0} & \textbf{115.0} & \textbf{130.7} & \textbf{141.8} & \textbf{71.8} & \textbf{89.9} & \textbf{104.9} & \textbf{115.2}\\\hline
& \multicolumn{4}{c}{Waiting} & \multicolumn{4}{c}{Walking Dog}&\multicolumn{4}{c}{Walking Together}&\multicolumn{4}{c}{Average} \\
milliseconds & 560 & 720 & 880 & 1000 & 560 & 720 & 880 & 1000 & 560 & 720 & 880 & 1000 & 560 & 720 & 880 & 1000\\\hline
Res. Sup.~\citep{Martinez_2017_CVPR} & 105.4 & 117.3 & 128.1 & 135.4 & 128.7 & 141.1 & 155.3 & 164.5 & 80.2 & 87.3 & 92.8 & 98.2 & 106.3 & 119.4 & 130.0 & 136.6 \\
convSeq2Seq~\citep{LiZLL18} & 87.3 & 100.3 & 110.7 & 117.7 & 122.4 & 133.8 & 151.1 & 162.4 & 72.0 & 77.7 & 82.9 & 87.4 & 90.7 & 104.7 & 116.7 & 124.2 \\
LTD-50-25\citep{mao2019learning} & 77.2 & 90.6 & 101.1 & 108.3 & 107.8 & 120.3 & 136.3 & 146.4 & 56.0 & 60.3 & 63.1 & 65.7 & 79.6 & 93.6 & 105.2 & 112.4 \\
LTD-10-25\citep{mao2019learning} & 75.1 & 88.7 & 99.5 & 106.9 & 105.8 & 118.7 & 132.8 & 142.2 & 58.0 & 63.6 & 67.0 & 69.6 & 79.5 & 94.0 & 105.6 & 112.7 \\
LTD-10-10\citep{mao2019learning} & 73.4 & 88.2 & 99.8 & 107.5 & 109.7 & 122.8 & 139.0 & 150.1 & 55.7 & 61.3 & 66.4 & 69.8 & 78.3 & 93.3 & 106.0 & 114.0 \\\hline
Pose Motion Att.~\citep{mao2020history} & 74.5 & 89.0 & 100.3 & 108.2 & 108.2 & 120.6 & 135.9 & 146.9 & 52.7 & 57.8 & 62.0 & 64.9 & 77.3 & 91.8 & 104.1 & 112.1 \\
Motion Att. + Post-fusion & \textbf{72.7} & \textbf{86.9} & \textbf{97.6} & \textbf{105.1} & \textbf{105.1} & \textbf{117.5} & \textbf{131.6} & \textbf{141.4} & \textbf{51.2} & \textbf{56.2} & \textbf{60.3} & \textbf{63.2} & \textbf{75.9} & \textbf{90.4} & \textbf{102.5} & \textbf{110.1}\\\hline
\end{tabu}
}
\caption{Long-term prediction of 3D joint positions on H3.6M. On average, our approach performs the best. Note that, on ``Walking" at $1000ms$, the 3D error of our method is $17\%$ lower than that of LTD-10-10~\citep{mao2019learning}, which uses the same predictor but no attention model.}
\label{tab:h36_long_3d}
\end{table*}

\subsection{Training}\label{sec:train}
Let us now introduce the loss functions we use to train our model on either 3D coordinates or joint angles.
For 3D joint coordinates prediction, 
we make use of the Mean Per Joint Position Error (MPJPE) proposed in~\citep{h36m_pami}. In particular, for one training sample, this yields the loss
\begin{equation}
    \ell = \frac{1}{J(M+T)}\sum_{t=1}^{M+T}\sum_{j=1}^{J}\|\hat{\textbf{p}}_{t,j}-\textbf{p}_{t,j}\|^2\;,
\end{equation}
where $\hat{\textbf{p}}_{t,j}\in \mathbf{R}^3$ represents the 3D coordinates of the $j^{th}$ joint of the $t^{th}$ human pose in $\hat{{\bf X}}_{N-M+1:N+T}$, and $\textbf{p}_{t,j}\in \mathbf{R}^3$ is the corresponding ground truth.

For the angle-based representation, we use the average $\ell_1$ distance between the predicted joint angles and the ground truth as loss. For one sample, this can be expressed as
\begin{equation}
    \ell = \frac{1}{K(M+T)}\sum_{t=1}^{M+T}\sum_{k=1}^{K}|\hat{x}_{t,k}-x_{t,k}|\;,
\end{equation}
where $\hat{x}_{t,k}$ is the predicted $k^{th}$ angle of the $t^{th}$ pose in $\hat{{\bf X}}_{N-M+1:N+T}$ and $x_{t,k}$ is the corresponding ground truth.
\begin{figure*}[ht]
    \centering
    \begin{tabular}{cc}
      \includegraphics[width=0.46\linewidth]{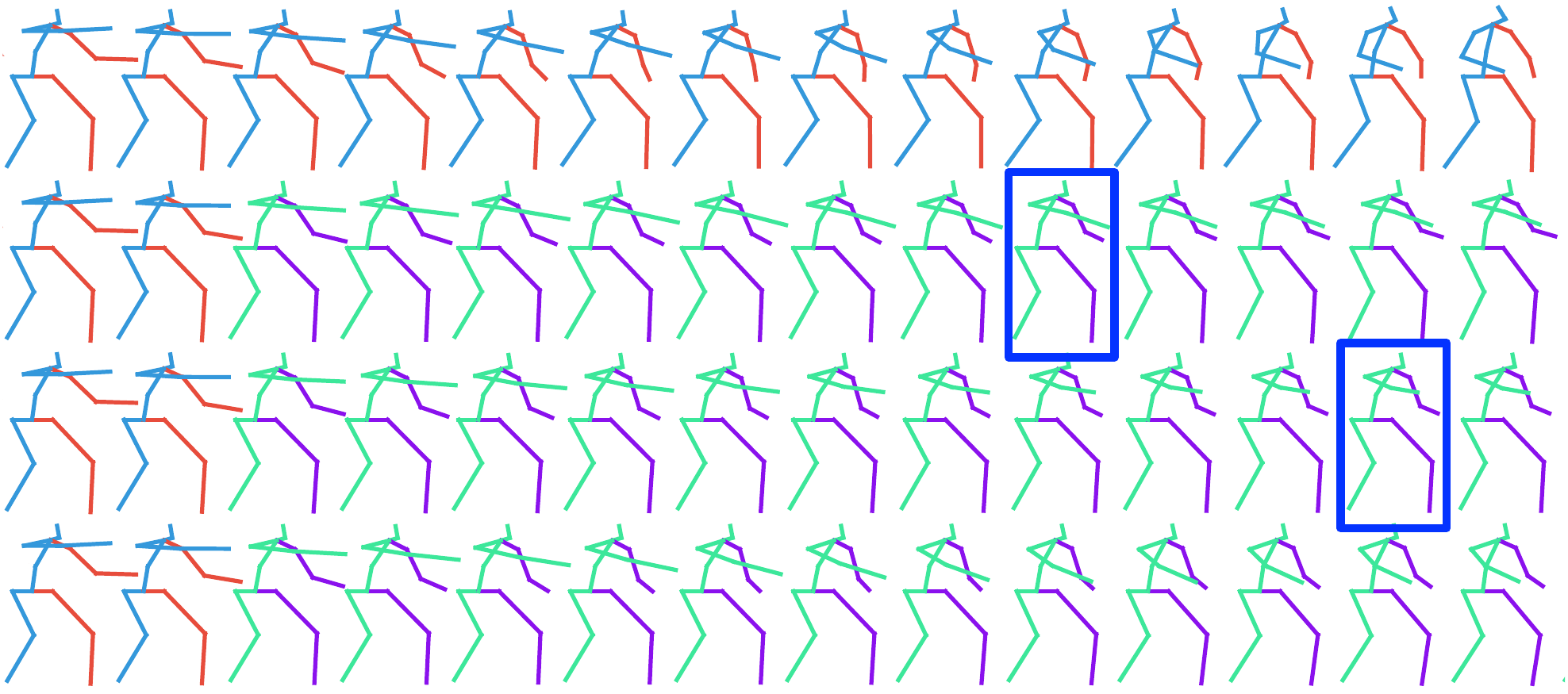} & 
      \includegraphics[width=0.46\linewidth]{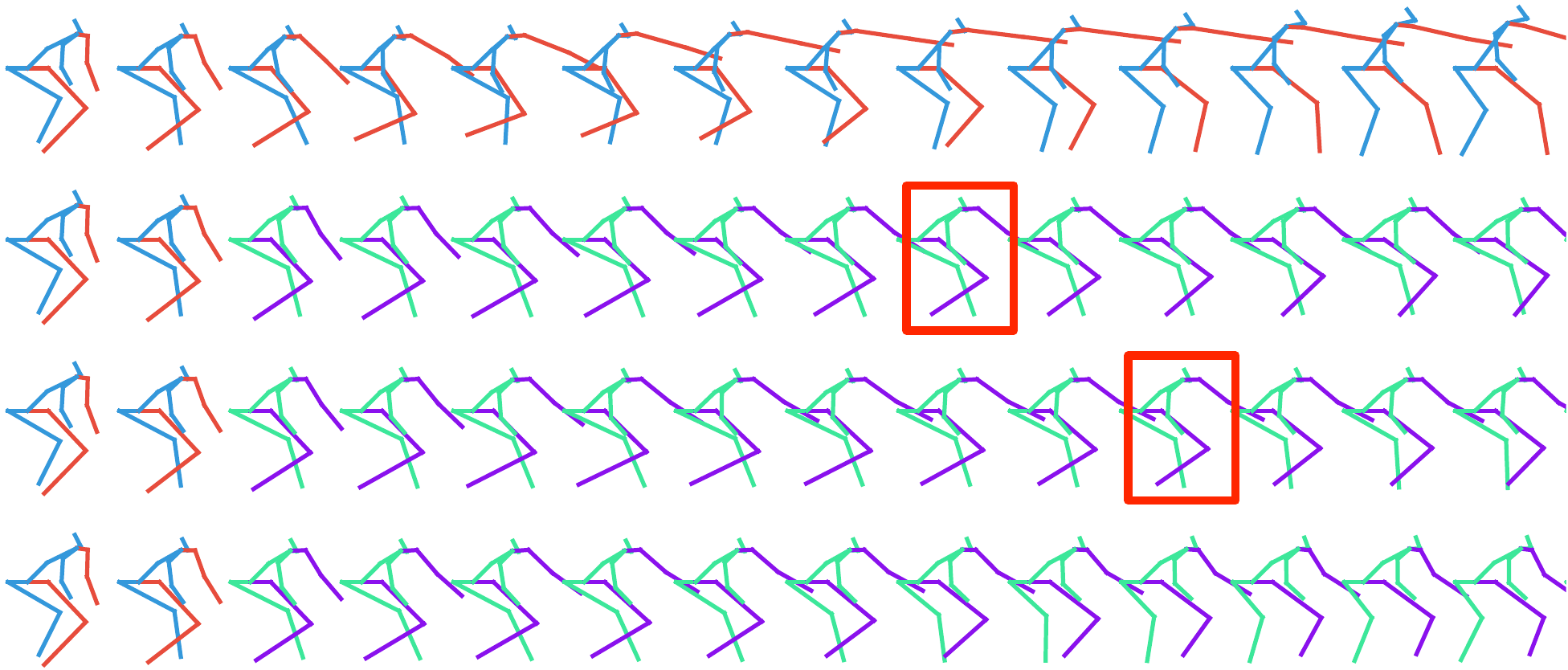} \\
      (a)Discussion & (b)Walking Dog\\
      \multicolumn{2}{c}{\includegraphics[width=0.96\linewidth]{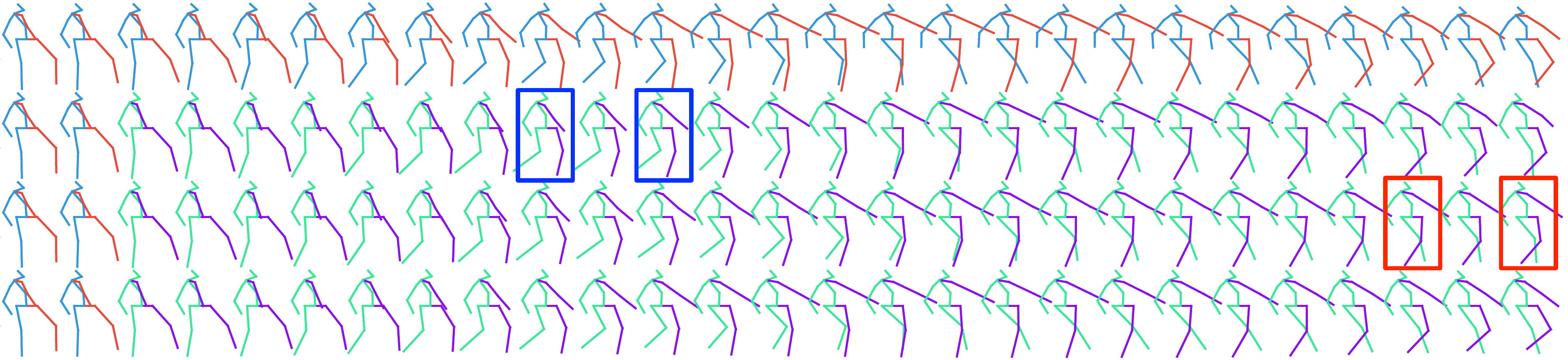}}\\
      \multicolumn{2}{c}{(c)Walking}
    \end{tabular}
    \caption{Qualitative comparison of short-term (``Discussion" and ``Walking Dog") and long-term (``Walking") predictions on H3.6M. From top to bottom, we show the ground truth, and the results of LTD-10-25, LTD-10-10 and our approach on 3D positions. The ground truth is shown as blue-red skeletons, and the predictions as green-purple ones.} 
    \label{fig:short-pred-quali-h36m}
\end{figure*}
\subsection{Network Structure}
As shown in Fig.~\ref{fig:net-structure}, our motion prediction framework consists of two modules: a motion attention model and a predictor. For the attention model, we use the same architecture for $f_q^p$ and $f_k^p$. Specifically, we use a network consisting of two 1D convolutional layers, each of which is followed by a ReLU activation function. In our experiments, the kernel size of these two layers is 6 and 5, respectively, to obtain a receptive field of 10 frames. The dimension of the hidden features, the query vector $\textbf{q}^p$ and the key vectors $\{\textbf{k}_i^p\}_{i=1}^{N-M-T+1}$ is set to 256. 

For the predictor, we use the same GCN with residual structure as in our previous work~\citep{mao2019learning}. It is made of 12 residual blocks, each of which contains two graph convolutional layers, with an additional initial layer to map the DCT coefficients to features and a final layer to decode the features to DCT residuals. Details of the predictor network structure are shown in Fig.~\ref{fig:pred_model}. The learnable weight matrix $\textbf{W}$ of each layer is of size $256\times 256$, and the size of the learnable adjacency matrix $\textbf{A}$ depends on the dimension of one human pose. For example, for 3D coordinates, $\textbf{A}$ is of size $66\times66$. Thanks to the simple structure of our attention model, the overall network remains still compact. Specifically, in our experiments, it has around 3.4 million parameters for both 3D coordinates and angles. The implementation details are included in supplementary material.
For the fusion model, we use a similar GCN-based network structure as the predictor but without the overall residual connection. 
\section{Experiments}\label{sec:exps}
Following previous works~\citep{gopalakrishnan2019neural,LiZLL18,mao2019learning,Martinez_2017_CVPR,pavllo2019modeling}, we evaluate our method on Human3.6m (H3.6M)~\citep{h36m_pami} and AMASS~\citep{AMASS:2019}. We further evaluate our method on 3DPW~\citep{vonMarcard2018} using our model trained on AMASS to demonstrate the generalizability of our approach. Below, we discuss these datasets, the evaluation metric and the baseline methods, and present our results using joint angles and 3D coordinates.
\begin{table*}[ht]
\centering
\resizebox{0.9\textwidth}{!}{%
\begin{tabular}{ccccc|cccc|cccc|cccc}
  & \multicolumn{4}{c}{Walking} & \multicolumn{4}{c}{Eating} & \multicolumn{4}{c}{Smoking} & \multicolumn{4}{c}{Discussion} \\
     milliseconds       & 80   & 160  & 320  & 400  & 80   & 160  & 320  & 400  & 80   & 160  & 320  & 400  & 80   & 160  & 320  & 400   \\\hline
Res. sup.~\citep{Martinez_2017_CVPR}& 0.36 & 0.63 & 0.95 & 1.04 & 0.30 & 0.58 & 0.90 & 1.01 & 0.37 & 0.73 & 1.19 & 1.33 & 0.56 & 0.90 & 1.37 & 1.51 \\
convSeq2Seq~\citep{LiZLL18} &  0.28 & 0.47 & 0.68 & 0.75 & 0.21 & 0.35 & 0.57 & 0.71 & 0.29 & 0.50 & 0.94 & 0.90 & 0.31 & 0.63 & 0.89 & 1.01 \\
LTD-10-25~\citep{mao2019learning} & 0.26 & 0.47 & 0.73 & 0.80 & 0.21 & 0.45 & 0.71 & 0.82 & 0.26 & 0.43 & 0.74 & 0.86 & 0.48 & 0.67 & 1.10 & 1.28\\
LTD-10-10~\citep{mao2019learning} & 0.25 & 0.45 & 0.72 & 0.78 & 0.20 & 0.41 & 0.70 & 0.82 & 0.25 & 0.41 & 0.71 & 0.83 & 0.47 & 0.68 & 1.09 & 1.25\\
QuaterNet vel.~\citep{pavllo2019modeling} & 0.28 & 0.49 & 0.76 & 0.83 & 0.22 & 0.47 & 0.76 & 0.88 & 0.28 & 0.47 & 0.79 & 0.91 & 0.48 & 0.74 & 1.20 & 1.37 \\\hline
Pose Motion Att.~\citep{mao2020history}  & \textbf{0.24} & \textbf{0.43} & \textbf{0.66} & \textbf{0.71} & \textbf{0.20} & \textbf{0.41} & \textbf{0.68} & \textbf{0.80} & \textbf{0.25} & \textbf{0.41} & \textbf{0.71} & \textbf{0.83} & \textbf{0.44} & \textbf{0.68} & \textbf{1.09} & \textbf{1.25} \\
Motion Att. + Post-fusion  & \textbf{0.24} & \textbf{0.43} & \textbf{0.66} & \textbf{0.71} & \textbf{0.20} & \textbf{0.41} & 0.69 & \textbf{0.80} & \textbf{0.25} & \textbf{0.41} & 0.74 & 0.86 & \textbf{0.44} & \textbf{0.68} & \textbf{1.09} & 1.26\\\hline\hline
MHU (8 Sub-seq)~\citep{Tang_2018} & 0.32 & 0.53 & 0.69 & 0.77 & - & - & - & - & - & - & - & - & 0.31 & 0.66 & 0.93 & 1.00 \\
LJP (8 Sub-seq)~\citep{cai2020learning} 
 & \textbf{0.17} & \textbf{0.30} & 0.51 & 0.55 & \textbf{0.16} & 0.29 & 0.50 & 0.61 & \textbf{0.21} & \textbf{0.40} & \textbf{0.85} & \textbf{0.78} & \textbf{0.19} & 0.54 & 0.89 & 0.94\\\hline
Motion Att. + Post-fusion (8 Sub-seq) & 0.18 & 0.31 & \textbf{0.48} & \textbf{0.52} & \textbf{0.16} & \textbf{0.28} & \textbf{0.47} & \textbf{0.59} & 0.22 & 0.41 & 0.86 & 0.80 & \textbf{0.19} & \textbf{0.51} & \textbf{0.77} & \textbf{0.85}\\\hline
& \multicolumn{4}{c}{Directions} & \multicolumn{4}{c}{Greeting} & \multicolumn{4}{c}{Phoning} & \multicolumn{4}{c}{Posing}\\
milliseconds & 80   & 160  & 320  & 400  & 80   & 160  & 320  & 400  & 80   & 160  & 320  & 400  & 80   & 160  & 320  & 400   \\\hline
Res. sup.~\citep{Martinez_2017_CVPR}& 0.33 & 0.59 & 0.98 & 1.15 & 0.66 & 1.00 & 1.44 & 1.59 & 0.44 & 0.73 & 1.15 & 1.31 & 0.51 & 0.88 & 1.39 & 1.59 \\
convSeq2Seq~\citep{LiZLL18} & 0.42 & 0.65 & 0.81 & 0.91 & 0.52 & 0.87 & 1.27 & 1.43 & 0.59 & 1.14 & 1.51 & 1.64 & 0.33 & 0.65 & 1.24 & 1.52 \\
LTD-10-25~\citep{mao2019learning} & 0.20 & 0.41 & 0.76 & 0.92 & 0.52 & 0.84 & 1.24 & 1.41 & 0.34 & 0.57 & 0.96 & 1.09 & 0.31 & 0.60 & 1.06 & 1.24 \\
LTD-10-10~\citep{mao2019learning} & \textbf{0.19} & 0.39 & 0.75 & 0.91 & 0.53 & 0.82 & 1.22 & 1.39 & 0 .33 & \textbf{0.54} & 0.94 & 1.07 & 0.30 & 0.61 & 1.02 & \textbf{1.20} \\
QuaterNet vel.~\citep{pavllo2019modeling} & 0.24 & 0.46 & 0.84 & 1.01 & 0.61 & 0.93 & 1.34 & 1.51 & 0.36 & 0.61 & 0.98 & 1.14 & 0.38 & 0.71 & 1.20 & 1.39 \\\hline
Pose Motion Att.~\citep{mao2020history} & \textbf{0.19} & \textbf{0.38} & 0.74 & 0.90 & \textbf{0.50} & \textbf{0.79} & 1.21 & 1.38 & \textbf{0.32} & \textbf{0.54} & 0.94 & 1.07 & \textbf{0.27} & \textbf{0.57} & \textbf{1.00} & 1.22\\
Motion Att. + Post-fusion & \textbf{0.19} & \textbf{0.38} & \textbf{0.73} & \textbf{0.89} & \textbf{0.50} & 0.81 & \textbf{1.19} & \textbf{1.36} & \textbf{0.32} & \textbf{0.54} & \textbf{0.93} & \textbf{1.06} & 0.29 & \textbf{0.57} & 1.01 & 1.21\\\hline\hline
MHU (8 Sub-seq)~\citep{Tang_2018} & - & - & - & - & 0.54 & 0.87 & 1.27 & 1.45 & - & - & - & - & 0.33 & 0.64 & 1.22 & 1.47 \\
LJP (8 Sub-seq)~\citep{cai2020learning} & \textbf{0.22} & \textbf{0.39} & \textbf{0.62} & \textbf{0.69} & \textbf{0.34} & \textbf{0.58} & 0.94 & 1.12 & \textbf{0.46} & \textbf{0.90} & \textbf{1.20} & \textbf{1.37} & \textbf{0.19} & \textbf{0.44} & 1.07 & 1.30\\\hline
Motion Att. + Post-fusion (8 Sub-seq) & 0.25 & 0.42 & 0.63 & 0.72 & 0.35 & 0.59 & \textbf{0.92} & \textbf{1.10} & 0.53 & 1.01 & 1.31 & 1.43 & 0.20 & \textbf{0.44} & \textbf{1.03} & \textbf{1.28}\\\hline
&\multicolumn{4}{c}{Purchases}&\multicolumn{4}{c}{Sitting} & \multicolumn{4}{c}{Sitting Down} & \multicolumn{4}{c}{Taking Photo} \\
milliseconds & 80   & 160  & 320  & 400  & 80   & 160  & 320  & 400  & 80   & 160  & 320  & 400  & 80   & 160  & 320  & 400   \\\hline
Res. sup.~\citep{Martinez_2017_CVPR}& 0.58 & 0.98 & 1.37 & 1.47 & 0.44 & 0.81 & 1.29 & 1.46 & 0.62 & 1.07 & 1.65 & 1.85 & 0.32 & 0.56 & 0.94 & 1.09 \\
convSeq2Seq~\citep{LiZLL18} & 0.62 & 0.89 & 1.18 & 1.25 & 0.41 & 0.64 & 1.03 & 1.20 & 0.41 & 0.76 & \textbf{1.13} & \textbf{1.26} & 0.29 & 0.52 & 0.81 & 0.95 \\
LTD-10-25~\citep{mao2019learning} & 0.47 & 0.84 & 1.24 & 1.33 & 0.33 & 0.52 & 0.92 & 1.06 & 0.44 & 0.75 & 1.21 & 1.40 & 0.21 & 0.35 & 0.62 & 0.74\\
LTD-10-10~\citep{mao2019learning} & 0.45 & 0.80 & 1.22 & 1.32 & 0.28 & 0.56 & \textbf{0.94} & 1.08 & \textbf{0.43} & 0.74 & 1.20 & 1.38 & 0.20 & \textbf{0.34} & 0.61 & 0.72 \\
QuaterNet vel.~\citep{pavllo2019modeling} & 0.54 & 0.92 & 1.36 & 1.47 & 0.34 & 0.59 & 1.00 & 1.15 & 0.47 & 0.81 & 1.31 & 1.50 & 0.23 & 0.39 & 0.69 & 0.81\\\hline
Pose Motion Att.~\citep{mao2020history} & \textbf{0.43} & 0.79 & 1.21 & 1.32 & \textbf{0.27} & \textbf{0.56} & \textbf{0.94} & 1.06 & \textbf{0.43} & 0.74 & 1.20 & 1.39 & \textbf{0.19} & \textbf{0.34} & \textbf{0.60} & 0.72\\
Motion Att. + Post-fusion & \textbf{0.43} & \textbf{0.78} & \textbf{1.20} & \textbf{1.30} & \textbf{0.27} & \textbf{0.56} & 0.96 & \textbf{1.05} & \textbf{0.43} & \textbf{0.73} & 1.19 & 1.38 & \textbf{0.19} & \textbf{0.34} & \textbf{0.60} & \textbf{0.71} \\\hline\hline
MHU (8 Sub-seq)~\citep{Tang_2018} & - & - & - & - & 0.27 & 0.54 & 0.84 & 0.96 & - & - & - & - & - & - & - & - \\
LJP (8 Sub-seq)~\citep{cai2020learning} & \textbf{0.38} & \textbf{0.64} & 1.13 & 1.21 & \textbf{0.27} & \textbf{0.44} & \textbf{0.78} & \textbf{0.96} & \textbf{0.27} & \textbf{0.54} & \textbf{0.88} & \textbf{0.97} & \textbf{0.13} & \textbf{0.33} & 0.60 & 0.74\\\hline
Motion Att. + Post-fusion (8 Sub-seq) & 0.44 & 0.65 & \textbf{1.00} & \textbf{1.06} & 0.29 & 0.46 & 0.81 & 0.99 & 0.30 & 0.63 & 0.91 & 1.02 & 0.15 & 0.35 & \textbf{0.57} & \textbf{0.69}\\\hline
& \multicolumn{4}{c}{Waiting} & \multicolumn{4}{c}{Walking Dog}&\multicolumn{4}{c}{Walking Together}&\multicolumn{4}{c}{Average} \\
milliseconds & 80   & 160  & 320  & 400  & 80   & 160  & 320  & 400  & 80   & 160  & 320  & 400  & 80   & 160  & 320  & 400   \\\hline

Res. sup.~\citep{Martinez_2017_CVPR}& 0.44 & 0.74 & 1.27 & 1.46 & 0.53 & 0.85 & 1.22 & 1.33 & 0.36 & 0.59 & 0.87 & 0.99 & 0.45 & 0.78 & 1.20 & 1.35 \\
convSeq2Seq~\citep{LiZLL18} & 0.33 & 0.65 & 1.13 & 1.33 & 0.58 & 0.97 & 1.36 & 1.49 & 0.28 & 0.54 & 0.72 & 0.75 & 0.39 & 0.68 & 1.02 & 1.14\\
LTD-10-25~\citep{mao2019learning} & 0.29 & 0.49 & 0.92 & 1.07 & 0.44 & 0.71 & 1.04 & 1.14 & 0.26 & 0.43 & 0.67 & 0.77 & 0.34 & 0.57 & 0.93 & 1.06 \\
LTD-10-10~\citep{mao2019learning} & 0.28 & \textbf{0.47} & \textbf{0.90} & \textbf{1.05} & 0.43 & 0.69 & 1.02 & 1.13 & \textbf{0.24} & 0.40 & 0.63 & 0.73 & 0.32 & \textbf{0.55} & 0.91 & 1.04 \\
QuaterNet vel.~\citep{pavllo2019modeling} & 0.32 & 0.54 & 1.00 & 1.15 & 0.48 & 0.78 & 1.12 & 1.21 & 0.28 & 0.45 & 0.69 & 0.79 & 0.37 & 0.62 & 1.00 & 1.14\\\hline
Pose Motion Att.~\citep{mao2020history} & \textbf{0.27} & \textbf{0.47} & 0.91 & 1.07 & \textbf{0.42} & \textbf{0.68} & 1.01 & 1.12 & \textbf{0.24} & \textbf{0.39} & 0.62 & \textbf{0.71} & \textbf{0.31} & \textbf{0.55} & \textbf{0.90} & 1.04\\
Motion Att. + Post-fusion & \textbf{0.27} & \textbf{0.47} & 0.91 & 1.06 & \textbf{0.42} & \textbf{0.68} & \textbf{1.00} & \textbf{1.11} & \textbf{0.24} & \textbf{0.39} & \textbf{0.61} & \textbf{0.71} & \textbf{0.31} & \textbf{0.55} & \textbf{0.90} & \textbf{1.03}\\\hline\hline
MHU (8 Sub-seq)~\citep{Tang_2018} & - & - & - & - & 0.56 & 0.88 & 1.21 & 1.37 & - & - & - & - & 0.39 & 0.68 & 1.01 & 1.13 \\
LJP (8 Sub-seq)~\citep{cai2020learning} & \textbf{0.21} & \textbf{0.48} & \textbf{0.86} & \textbf{1.08} & \textbf{0.40} & \textbf{0.75} & \textbf{1.05} & 1.23 & \textbf{0.14} & \textbf{0.32} & 0.52 & \textbf{0.55} & \textbf{0.25} & \textbf{0.49} & 0.83 & 0.94\\\hline
Motion Att. + Post-fusion (8 Sub-seq) & 0.23 & 0.49 & 0.90 & 1.11 & 0.46 & 0.77 & \textbf{1.05} & \textbf{1.21} & \textbf{0.14} & \textbf{0.32} & \textbf{0.50} & \textbf{0.55} & 0.27 & 0.51 & \textbf{0.81} & \textbf{0.93}\\\hline
\end{tabular}
}
\caption{Short-term prediction of joint angles on H3.6M. Following QuaterNet~\citep{pavllo2019modeling}, we report the average error on 256 sub-sequences, except when indicating ``(8 Sub-seq)" after a method's name, in which case the error is averaged over 8 sub-sequences per action, as reported in the corresponding paper.}
\label{tab:h36_short_ang}
\end{table*}
\subsection{Datasets}
\noindent{{\bf Human3.6M}}~\citep{h36m_pami} is the most widely used benchmark dataset for motion prediction. It depicts seven actors performing 15 actions. Each human pose is represented as a 32-joint skeleton. We compute the 3D coordinates of the joints by applying forward kinematics on a standard skeleton as in~\citep{mao2019learning}. Following~\citep{LiZLL18,mao2019learning,Martinez_2017_CVPR}, we remove the global rotation, translation and constant angles or 3D coordinates of each human pose, and down-sample the motion sequences to 25 frames per second. As previous work~\citep{LiZLL18,mao2019learning,Martinez_2017_CVPR}, we test our method on subject 5 (S5). However, instead of testing on only 8 random sub-sequences per action, which was shown in~\citep{pavllo2019modeling} to lead to high variance, we report our results on 256 sub-sequences per action. Nevertheless, for the baselines~\citep{Tang_2018,cai2020learning} whose code is not publicly available, we compare our results to theirs on the same 8 sub-sequences of each action.

\noindent{{\bf AMASS}}. The Archive of Motion Capture as Surface Shapes (AMASS) dataset~\citep{AMASS:2019} is a recently published human motion dataset, which unifies many mocap datasets, such as CMU, KIT and BMLrub, using a SMPL~\citep{SMPL:2015,MANO:SIGGRAPHASIA:2017} parameterization to obtain a human mesh. SMPL represents a human by a shape vector and joint rotation angles. The shape vector, which encompasses coefficients of different human shape bases, defines the human skeleton. We obtain human poses in 3D by applying forward kinematics to one human skeleton. In AMASS, a human pose is represented by 52 joints, including 22 body joints and 30 hand joints. Since we focus on predicting human body motion, we discard the hand joints and the 4 static joints, leading to an 18-joint human pose. As for H3.6M, we down-sample the frame-rate to 25Hz.
\begin{table*}[ht]
\centering
\resizebox{0.9\textwidth}{!}{%
\begin{tabular}{ccccc|cccc|cccc|cccc}
  & \multicolumn{4}{c}{Walking} & \multicolumn{4}{c}{Eating} & \multicolumn{4}{c}{Smoking} & \multicolumn{4}{c}{Discussion} \\
 milliseconds & 560 & 720 & 880 & 1000 & 560 & 720 & 880 & 1000 & 560 & 720 & 880 & 1000 & 560 & 720 & 880 & 1000\\\hline

Res. sup.~\citep{Martinez_2017_CVPR}& 1.21 & 1.32 & 1.41 & 1.47 & 1.19 & 1.36 & 1.47 & 1.55 & 1.44 & 1.57 & 1.68 & 1.76 & 1.78 & 1.92 & 2.04 & 2.12\\
convSeq2Seq~\citep{LiZLL18} & 0.87 & 0.96 & 0.97 & 1.00 & 0.86 & 0.90 & 1.12 & 1.24 & 0.98 & 1.11 & 1.42 & 1.67 & 1.42 & 1.76 & 1.90 & 2.03\\
LTD-10-25~\citep{mao2019learning} & 0.92 & 0.97 & 1.03 & 1.05 & 0.99 & 1.16 & 1.26 & 1.33 & 1.07 & 1.26 & 1.41 & 1.55 & 1.48 & 1.59 & \textbf{1.68} & \textbf{1.76}\\
LTD-10-10~\citep{mao2019learning} 0.95 & 1.03 & 1.09 & 1.12 & 0.98 & 1.15 & 1.28 & 1.36 & \textbf{1.04} & 1.21 & \textbf{1.36} & 1.51 & \textbf{1.47} & 1.59 & 1.71 & 1.79  \\\hline
Pose Motion Att.~\citep{mao2020history} & \textbf{0.84} & \textbf{0.91} & \textbf{0.99} & \textbf{1.03} & \textbf{0.98} & 1.14 & 1.24 & 1.31 & \textbf{1.04} & \textbf{1.20} & 1.38 & 1.50 & 1.49 & 1.62 & 1.72 & 1.82\\
Motion Att. + Post-fusion & 0.85 & \textbf{0.91} & \textbf{0.99} & \textbf{1.03} & \textbf{0.98} & \textbf{1.13} & \textbf{1.23} & \textbf{1.30} & 1.07 & 1.23 & 1.38 & \textbf{1.47} & 1.50 & \textbf{1.57} & 1.72 & 1.79\\\hline\hline
MHU (8 Sub-seq)~\citep{Tang_2018} & 1.44 & 1.46 & - & 1.44 & - & - & - & - & - & - & - & - & 1.37 & 1.66 & - & 1.88 \\
Motion Att. + Post-fusion (8 Sub-seq)& \textbf{0.58} & \textbf{0.62} & 0.61 & \textbf{0.63} & 0.73 & 0.80 & 0.99 & 1.09 & 0.86 & 1.00 & 1.34 & 1.57 & \textbf{1.27} & \textbf{1.52} & 1.65 & \textbf{1.71}\\\hline

& \multicolumn{4}{c}{Directions} & \multicolumn{4}{c}{Greeting} & \multicolumn{4}{c}{Phoning} & \multicolumn{4}{c}{Posing}\\
 milliseconds & 560 & 720 & 880 & 1000 & 560 & 720 & 880 & 1000 & 560 & 720 & 880 & 1000 & 560 & 720 & 880 & 1000\\\hline

Res. sup.~\citep{Martinez_2017_CVPR}& 1.35 & 1.50 & 1.63 & 1.72 & 1.82 & 2.02 & 2.16 & 2.21 & 1.52 & 1.70 & 1.85 & 1.96 & 1.90 & 2.13 & 2.37 & 2.46\\
convSeq2Seq~\citep{LiZLL18} & 1.00 & 1.18 & 1.41 & 1.44 & 1.73 & 1.75 & 1.92 & 1.90 & 1.66 & 1.81 & 1.93 & 2.05 & 1.95 & 2.26 & 2.49 & 2.63\\
LTD-10-25~\citep{mao2019learning} & 1.10 & 1.23 & 1.35 & \textbf{1.41} & 1.63 & 1.81 & 1.95 & 2.01 & 1.29 & 1.48 & 1.63 & 1.74 & 1.54 & 1.81 & 2.10 & 2.23\\
LTD-10-10~\citep{mao2019learning} & 1.09 & \textbf{1.21} & \textbf{1.34} & \textbf{1.41} & 1.63 & 1.82 & 1.99 & 2.06 & 1.29 & 1.50 & 1.67 & 1.78 & \textbf{1.53} & 1.81 & 2.12 & 2.25 \\\hline
Pose Motion Att.~\citep{mao2020history} & \textbf{1.08} & 1.22 & 1.35 & 1.42 & 1.62 & 1.79 & 1.93 & 1.99 & 1.28 & 1.49 & \textbf{1.65} & \textbf{1.76} & 1.55 & 1.80 & 2.10 & 2.24\\
Motion Att. + Post-fusion & \textbf{1.08} & \textbf{1.21} & \textbf{1.34} & \textbf{1.41} & \textbf{1.59} & \textbf{1.75} & \textbf{1.87} & \textbf{1.93} & \textbf{1.27} & \textbf{1.48} & \textbf{1.65} & \textbf{1.76} & \textbf{1.53} & \textbf{1.78} & \textbf{2.08} & \textbf{2.22}\\\hline\hline

MHU (8 Sub-seq)~\citep{Tang_2018} & - & - & - & - & 1.75 & 1.74 & - & 1.87 & - & - & - & - & 1.82 & 2.17 & - & 2.51 \\\hline
Motion Att. + Post-fusion (8 Sub-seq) & 0.83 & 1.02 & 1.25 & 1.30 & \textbf{1.45} & \textbf{1.46} & 1.60 & \textbf{1.56} & 1.41 & 1.56 & 1.67 & 1.67 & \textbf{1.54} & \textbf{1.83} & 2.15 & \textbf{2.33}\\\hline
&\multicolumn{4}{c}{Purchases}&\multicolumn{4}{c}{Sitting} & \multicolumn{4}{c}{Sitting Down} & \multicolumn{4}{c}{Taking Photo}\\
milliseconds & 560 & 720 & 880 & 1000 & 560 & 720 & 880 & 1000 & 560 & 720 & 880 & 1000 & 560 & 720 & 880 & 1000\\\hline

Res. sup.~\citep{Martinez_2017_CVPR}& 1.65 & 1.80 & 1.92 & 1.98 & 1.76 & 1.99 & 2.18 & 2.28 & 2.20 & 2.46 & 2.71 & 2.84 & 1.35 & 1.55 & 1.73 & 1.85\\
convSeq2Seq~\citep{LiZLL18} & 1.68 & 1.65 & 2.13 & 2.50 & 1.31 & 1.43 & 1.66 & 1.72 & 1.45 & 1.70 & 1.85 & 1.98 & 1.09 & 1.18 & 1.27 & \textbf{1.32}\\
LTD-10-25~\citep{mao2019learning} & 1.51 & 1.66 & 1.80 & 1.87 & 1.34 & 1.60 & 1.79 & 1.87&1.71 & 1.95 & 2.17 & 2.26 & 0.94 & 1.10 & 1.23 & 1.34\\
LTD-10-10~\citep{mao2019learning} & 1.52 & 1.68 & 1.83 & 1.91 & 1.34 & 1.60 & 1.79 & 1.89 &1.68 & 1.91 & 2.13 & 2.22 & 0.93 & 1.08 & 1.22 & 1.34\\\hline
Pose Motion Att.~\citep{mao2020history} & 1.47 & 1.62 & 1.75 & 1.82 & 1.33 & 1.59 & 1.79 & 1.88 & \textbf{1.68} & \textbf{1.90} & \textbf{2.12} & 2.22 & \textbf{0.92} & \textbf{1.07} & \textbf{1.21} & 1.33\\
Motion Att. + Post-fusion & \textbf{1.46} & \textbf{1.61} & \textbf{1.73} & \textbf{1.81} & \textbf{1.32} & \textbf{1.58} & \textbf{1.78} & \textbf{1.87} & \textbf{1.68} & \textbf{1.90} & \textbf{2.12} & \textbf{2.21} & \textbf{0.92} & \textbf{1.07} & \textbf{1.21} & \textbf{1.32}\\\hline\hline

MHU (8 Sub-seq)~\citep{Tang_2018} & - & - & - & - & \textbf{1.04} & \textbf{1.14} & - & \textbf{1.35} & - & - & - & - & - & - & - & - \\
Motion Att. + Post-fusion (8 Sub-seq)& 1.43 & 1.50 & 1.93 & 2.25 & 1.14 & 1.27 & 1.49 & 1.54 & 1.17 & 1.41 & 1.54 & 1.68 & 0.79 & 0.86 & 0.94 & 1.01\\\hline
& \multicolumn{4}{c}{Waiting} & \multicolumn{4}{c}{Walking Dog}&\multicolumn{4}{c}{Walking Together}&\multicolumn{4}{c}{Average} \\
milliseconds & 560 & 720 & 880 & 1000 & 560 & 720 & 880 & 1000 & 560 & 720 & 880 & 1000 & 560 & 720 & 880 & 1000\\\hline

Res. sup.~\citep{Martinez_2017_CVPR}& 1.74 & 1.95 & 2.13 & 2.25 & 1.50 & 1.68 & 1.77 & 1.86 & 1.18 & 1.30 & 1.38 & 1.45 & 1.57 & 1.75 & 1.90 & 1.99\\
convSeq2Seq~\citep{LiZLL18} & 1.68 & 2.02 & 2.33 & 2.45 & 1.73 & 1.85 & 1.99 & 2.04 & 0.82 & 0.89 & 0.95 & 1.29 & 1.35 & 1.50 & 1.69 & 1.82\\
LTD-10-25~\citep{mao2019learning} & \textbf{1.30} & 1.48 & \textbf{1.63} & \textbf{1.74} & 1.30 & 1.45 & 1.55 & 1.64 & 0.91 & 0.98 & 1.02 & 1.06 & 1.27 & 1.44 & 1.57 & 1.66\\
LTD-10-10~\citep{mao2019learning} & \textbf{1.30} & \textbf{1.47} & \textbf{1.63} & 1.75 & 1.31 & 1.48 & 1.59 & 1.68 & 0.89 & 0.98 & 1.03 & 1.08 & 1.26 & 1.44 & 1.59 & 1.68\\\hline
Pose Motion Att.~\citep{mao2020history} & 1.31 & 1.49 & 1.64 & 1.77 & 1.30 & 1.45 & 1.55 & 1.63 & \textbf{0.86} & 0.94 & 1.00 & \textbf{1.04} & \textbf{1.25} & 1.42 & 1.56 & 1.65\\
Motion Att. + Post-fusion & \textbf{1.30} & \textbf{1.47} & \textbf{1.63} & 1.75 & \textbf{1.28} & \textbf{1.44} & \textbf{1.54} & \textbf{1.62} & \textbf{0.86} & \textbf{0.93} & \textbf{0.99} & \textbf{1.04} & \textbf{1.25} & \textbf{1.40} & \textbf{1.55} & \textbf{1.64}\\\hline\hline

MHU (8 Sub-seq)~\citep{Tang_2018} & - & - & - & - & 1.67 & 1.81 & - & 1.90 & - & - & - & - & 1.34 & 1.49 & 1.69 & 1.80 \\\hline
Motion Att. + Post-fusion (8 Sub-seq)& 1.50 & 1.87 & 2.20 & 2.27 & \textbf{1.50} & \textbf{1.59} & 1.75 & \textbf{1.81} & 0.63 & 0.68 & 0.81 & 1.18 & \textbf{1.12} & \textbf{1.27} & \textbf{1.46} & \textbf{1.57}\\\hline
\end{tabular}
}
\caption{Long-term prediction of joint angles on H3.6M.}
\label{tab:h36_long_ang}
\end{table*}
Since most sequences of the official testing split\footnote[1]{Described at https://github.com/nghorbani/amass} of AMASS consist of transition between two irrelevant actions, such as dancing to kicking, kicking to pushing, they are not suitable to evaluate our prediction algorithms, which assume that the history is relevant to forecast the future. Therefore, instead of using this official split, we treat BMLrub\footnote[2]{Available at https://amass.is.tue.mpg.de/dataset.} (522 min. video sequence), as our test set as each sequence consists of one actor performing one type of action. We then split the remaining parts of AMASS into training and validation data.

\noindent{{\bf 3DPW}}. The 3D Pose in the Wild dataset (3DPW)~\citep{vonMarcard2018} consists of challenging indoor and outdoor actions. We only evaluate our model trained on AMASS on the test set of 3DPW to show the generalization of our approach.

As mentioned in Section~\ref{sec:att-model}, we model motion attention at 3 different levels. The full-pose and individual-joints ones are self-explanatory. For part-based motion attention,  we divide the human body into 5 parts following the human kinematic tree: torso (including neck and head), left arm, right arm, left leg and right leg. Each part consists of several human joints.
\begin{figure*}[ht]
    \centering
    \includegraphics[width=0.8\linewidth]{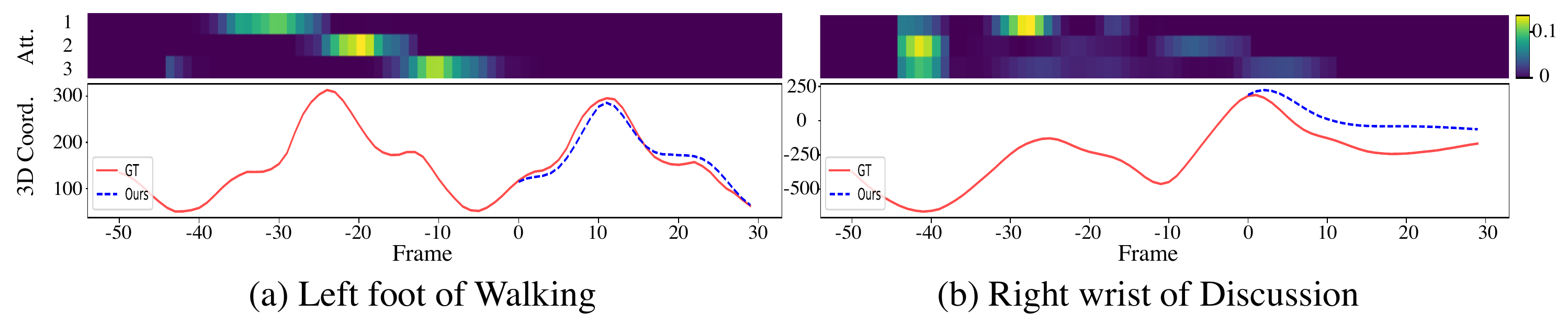}
    \caption{Visualization of attention maps and joint trajectories. The $x$-axis denotes the frame index, with prediction starting at frame 0. The $y$-axis of the attention map (top) is the prediction step. Specifically, since the model is trained to predict 10 future frames, we recursively perform prediction for 3 steps to generate 30 frames. Each row of an attention map is then the attention vector when predicting the corresponding 10 future frames. For illustration purpose, we show~\emph{per-frame} attention, which represents the attention for~\emph{its motion subsequence} consisting of M-1 frames forward and T frames afterwards. (a) Predicted attention map and trajectory of the left foot's $x$ coordinate for 'Walking', where the future motion closely resembles that between frames $-45$ and $-10$. Our model correctly attends to that very similar motion in the history. (b) Predicted attention map and trajectory of the right wrist's $x$ coordinate for 'Discussion'. In this case, the attention model searches for the most similar motion in the history. For example, in the $1^{st}$ prediction step, to predict frames $0$ to $10$ where a peak occurs, the model focuses on frames $-30$ to $-20$, where a similar peak pattern occurs.}
    \label{fig:att-h36m}
\end{figure*}
\begin{table*}[ht]
\centering
\resizebox{0.9\textwidth}{!}{%
\begin{tabular}{ccccccccc||cccccccc}
& \multicolumn{8}{c}{AMASS-BMLrub} & \multicolumn{8}{c}{3DPW} \\
milliseconds   & 80   & 160  & 320  & 400  & 560 & 720 & 880 & 1000 & 80   & 160  & 320  & 400  & 560 & 720 & 880 & 1000 \\\hline
convSeq2Seq~\citep{LiZLL18} & 20.6 & 36.9 & 59.7 & 67.6 & 79.0 & 87.0 & 91.5 & 93.5 & 18.8 & 32.9 & 52.0 & 58.8 & 69.4 & 77.0 & 83.6 & 87.8 \\
LTD-10-25~\citep{mao2019learning} & 11.0 & 20.7 & 37.8 & 45.3 & 57.2 & 65.7 & 71.3 & 75.2 & 12.6 & 23.2 & 39.7 & 46.6 & 57.9 & 65.8 & 71.5 & 75.5 \\
LTD-10-10~\citep{mao2019learning} & \textbf{10.3} & \textbf{19.3} & 36.6 & 44.6 & 61.5 & 75.9 & 86.2 & 91.2 & \textbf{12.0} & \textbf{22.0} & 38.9 & 46.2 & 59.1 & 69.1 & 76.5 & 81.1 \\\hline
Pose Motion Att.~\citep{mao2020history} & 11.3 & 20.7 & 35.7 & 42.0 & 51.7 & 58.6 & 63.4 & 67.2 & 12.6 & 23.1 & 39.0 & 45.4 & 56.0 & 63.6 & 69.7 & 73.7 \\
Motion Att. + Post-fusion & 11.0 & 20.3 & \textbf{35.0} & \textbf{41.2} & \textbf{50.7} & \textbf{57.4} & \textbf{61.9} & \textbf{65.8} & 12.4 & 22.6 & \textbf{38.1} & \textbf{44.4} & \textbf{54.7} & \textbf{62.1} & \textbf{67.9} & \textbf{71.8} \\\hline\hline
LTD-10-25~\citep{mao2019learning} & 0.21 & 0.37 & 0.62 & 0.71 & 0.83 & 0.91 & 0.95 & 0.98 & \textbf{0.38} & 0.64 & 1.00 & 1.16 & 1.34 & 1.50 & 1.60 & 1.69 \\\hline
Pose Motion Att.~\citep{mao2020history} & \textbf{0.20} & \textbf{0.36} & 0.57 & 0.65 & 0.75 & 0.83 & 0.87 & 0.92 & \textbf{0.38} & 0.64 & 0.99 & 1.15 & 1.32 & 1.49 & 1.59 & 1.68\\
Motion Att. + Post-fusion & \textbf{0.20} & \textbf{0.36} & \textbf{0.56} & \textbf{0.64} & \textbf{0.74} & \textbf{0.81} & \textbf{0.86} & \textbf{0.90} & \textbf{0.38} & \textbf{0.63} & \textbf{0.98} & \textbf{1.13} & \textbf{1.30} & \textbf{1.47} & \textbf{1.57} & \textbf{1.66} \\\hline
\end{tabular}
}
\caption{Short-term and long-term prediction of 3D joint positions (upper) and joint angles (bottom) on BMLrub (left) and 3DPW (right).}
\label{tab:amass_3d}
\end{table*}

\subsection{Evaluation Metrics and Baselines}
\noindent{{\bf Metrics}}. For the models that output 3D positions, we report the Mean Per Joint Position Error (MPJPE) \citep{h36m_pami} in millimeter, which is commonly used in human pose estimation. For those that predict angles, we follow the standard evaluation protocol~\citep{Martinez_2017_CVPR,LiZLL18,mao2019learning} and report the Euclidean distance in Euler angle representation. 

\noindent{{\bf Baselines.}} We compare our approach with two RNN-based methods, Res. sup.~\citep{Martinez_2017_CVPR} and MHU~\citep{Tang_2018}, two feed-forward models, convSeq2Seq~\citep{LiZLL18} and LTD~\citep{mao2019learning}, which constitutes the state of the art. We further compare it with the concurrent work LPJ~\citep{cai2020learning}, which exploits an attention-based transformer.
The angular results of Res. sup.~\citep{Martinez_2017_CVPR}, convSeq2Seq~\citep{LiZLL18} on H3.6M are obtained by running the official training code and report on 256 sub-sequences. For the other results of Res. sup.~\citep{Martinez_2017_CVPR} and convSeq2Seq~\citep{LiZLL18}, we adapt the code provided by the authors for H3.6M to 3D and AMASS. The results of MHU~\citep{Tang_2018} and LPJ~\citep{cai2020learning} on H3.6M are directly taken from the respective paper. For our LTD~\citep{mao2019learning}, we rely on the pre-trained models released for H3.6M, and train the model on AMASS using the released code.

While Res. sup.~\citep{Martinez_2017_CVPR}, convSeq2Seq \citep{LiZLL18} and MHU~\citep{Tang_2018} are all trained to generate 25 future frames, LTD \citep{mao2019learning} has 3 different models, which we refer to as LTD-50-25~\citep{mao2019learning}, LTD-10-25~\citep{mao2019learning}, and LTD-10-10~\citep{mao2019learning}. The two numbers after the method name indicate the number of observed past frames and that of future frames to predict, respectively, during training. For example, LTD-10-25~\citep{mao2019learning} means that the model is trained to take the past 10 frames as input to predict the future 25 frames.

\subsection{Results}\label{sec:exp-res}
Following the setting of our baselines~\citep{Martinez_2017_CVPR,LiZLL18,Tang_2018,mao2019learning}, we report results for short-term ($< 500ms$) and long-term ($> 500ms$) prediction. On H3.6M, our model is trained using the past $50$ frames to predict the future $10$ frames, and we produce poses further in the future by recursively applying the predictions as input to the model. On AMASS, our model is trained using the past $50$ frames to predict the future $25$ frames.

\noindent{{\bf Human3.6M.}} In Tables~\ref{tab:h36_short_3d} and~\ref{tab:h36_long_3d}, we provide the H3.6M results for short-term and long-term prediction in 3D space, respectively. Note that we outperform all the baselines on average for both short-term and long-term prediction. In particular, our method yields larger improvements on activities with a clear repeated history, such as ``Walking" and ``Walking Together". Nevertheless, our approach remains competitive on the other actions. Note that we consistently outperform LTD-50-25, which is trained on the same number of past frames as our approach. This, we believe, evidences the benefits of exploiting attention on the motion history. 

Moreover, the performance of our motion attention model (denoted as ``Pose Motion Att." in the tables) is consistently improved with the use of our fusion model, for both short-term and long-term prediction. Our approach performs comparable to the concurrent LPJ~\citep{cai2020learning}. Note that our motion attention strategy and fusion model are orthogonal to the progressive joint prediction and motion dictionary of LPJ~\citep{cai2020learning}, and thus one could expect further improvement by combining these two strategies. We further show that our methods with only the attention module outperforms that of LPJ~\citep{cai2020learning} by a large margin in Table~\ref{tab:h36_short_3d_no_prog}.

\begin{figure*}[!ht]
    \centering
      \includegraphics[width=\textwidth]{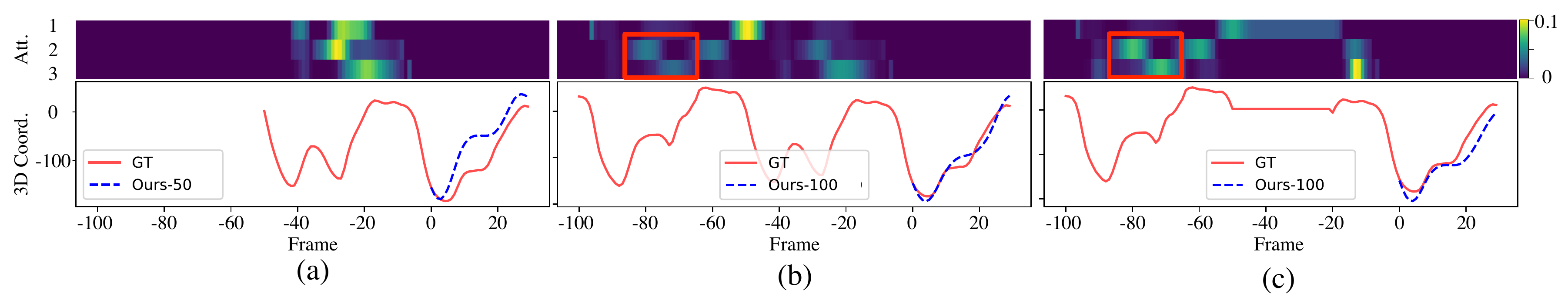}
      \caption{Visualization of attention maps and joint coordinate trajectories for ``Smoking" on H3.6M. (a) Results of our model observing 50 past frames. (b) Results of our model observing 100 frames. (c) Results obtained when replacing the motion of the past 40 frames with a constant pose.}
      \label{fig:his_rep_att}
\end{figure*}
Let us now focus on the LTD~\citep{mao2019learning} baseline, which constitutes the state of the art. Although LTD-10-10 is very competitive for short-term prediction, when it comes to generate poses in the further future, it yields higher average error, i.e., $114.0mm$ at $1000ms$. By contrast, LTD-10-25 and LTD-50-25 achieve good performance at $880ms$ and above, but perform worse than LTD-10-10 at other time horizons. Our approach with a unified model, however, yields state-of-the-art performance for both short-term and long-term predictions. To summarize, our motion attention model improves the performance of the predictor for short-term prediction and further enables it to generate better long-term predictions. This is further confirmed by Tables~\ref{tab:h36_short_ang} and~\ref{tab:h36_long_ang}, where we report the short-term and long-term prediction results in angle space on H3.6M, and by the qualitative comparison in Fig.~\ref{fig:short-pred-quali-h36m}. 

\begin{table}[!t]
\centering
\resizebox{0.5\textwidth}{!}{%
\begin{tabular}{ccccccccc}
milliseconds   & 80   & 160  & 320  & 400  & 560 & 720 & 880 & 1000 \\\hline
Ours-50 & \textbf{10.7} & \textbf{22.4} & 46.9 & 58.3 & 79.0 & 97.1 & 111.0 & 121.1\\
Ours-100 & \textbf{10.7} & 22.5 & \textbf{46.4} & \textbf{57.5} & \textbf{77.8} & \textbf{95.1} & \textbf{107.6} & \textbf{116.9}\\\hline
\end{tabular}
}
\caption{Short-term and long-term prediction of 3D positions on selected sequences where similar patterns occur in the longer history. The number after ``Ours" indicates the observed frames during testing. Both methods observed 50 frames during training.}
\label{tab:his_rep}
\end{table}
\noindent{{\bf AMASS \& 3DPW.}} The results of short-term and long-term prediction in 3D on AMASS and 3DPW are shown in Table~\ref{tab:amass_3d}. Our method consistently outperforms baseline approaches, which further shows the benefits of our motion attention model. Since none of the methods were trained on 3DPW, these results further demonstrate that our approach generalizes better to new datasets than the baselines.

\noindent\textbf{Visualisation of attention.} In Fig.~\ref{fig:att-h36m}, we visualize the attention maps computed by our motion attention model on a few sampled joints for their corresponding coordinate trajectories. In particular, we show attention maps for joints in a periodical motion (``Walking") and a non-periodical one (``Discussion"). In both cases, the attention model can find the most relevant sub-sequences in the history, which encode either a nearly identical motion (periodical action), or a similar pattern (non-periodical action).

\noindent{{\bf Motion repeats itself in longer-term history.}} Our model, which is trained with fixed-length observations, can nonetheless exploit longer history at test time if it is available. To evaluate this and our model's ability to capture long-range motion dependencies, we manually sampled $100$ sequences from the test set of H3.6M, in which similar motion occurs in the further past than that used to train our model.

In Table~\ref{tab:his_rep}, we compare the results of a model trained with 50 past frames and using either $50$ frames (Ours-50) or 100 frames (Ours-100) at test time. Although the performance is close in the very short term ($<160ms$), the benefits of our model using longer history become obvious when it comes to further future, leading to a performance boost of $4.2mm$ at $1s$. In Fig.~\ref{fig:his_rep_att}, we compare the attention maps and predicted joint trajectories of Ours-50 (a) and Ours-100 (b). The highlighted regions (in red box) in the attention map demonstrate that our model can capture the repeated motions in the further history if it is available during test and improve the motion prediction results.

To show the influence of further historical frames, we replace the past $40$ frames with a static pose, thus removing the motion in that period, and then perform prediction with this sequence. As shown in~Fig.~\ref{fig:his_rep_att} (c), attending to the similar motion between frames $-80$ and $-60$, yields a trajectory much closer to the ground truth than only attending to the past $50$ frames.

\noindent{{\bf Importance of different levels of attention.}} Our different levels of motion attention complement each other in the two ways discussed below. Note that, in this discussion, we categorize our 3 different levels of motion attention into 2 relative levels: global and local. For example, parts motion attention is referred to as {\it local-level attention} when compared to full pose motion attention, but as {\it global-level attention} when compared to joint motion attention.
\begin{figure*}[!ht]
    \centering
      \includegraphics[width=\linewidth]{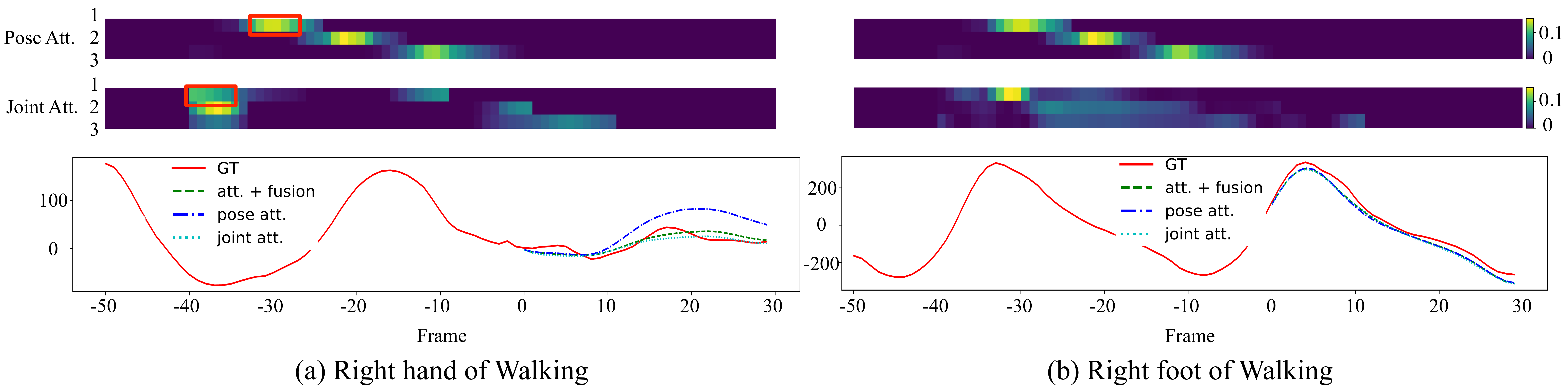}
      \caption{Visualization of attention maps and joint coordinate trajectories of different motion attentions for ``Walking'' on H3.6M. Pose motion attention captures the repeated motion for right foot (first row of (b)) while miss the the motion pattern for right hand (first row of (a)). Joint motion attention however, attends to the most relevant historical motion for both joints with two different attention maps and leads to a better prediction.}
      \label{fig:h36-att-walking}
\end{figure*} 
\begin{figure*}[!ht]
    \centering
      \includegraphics[width=\linewidth]{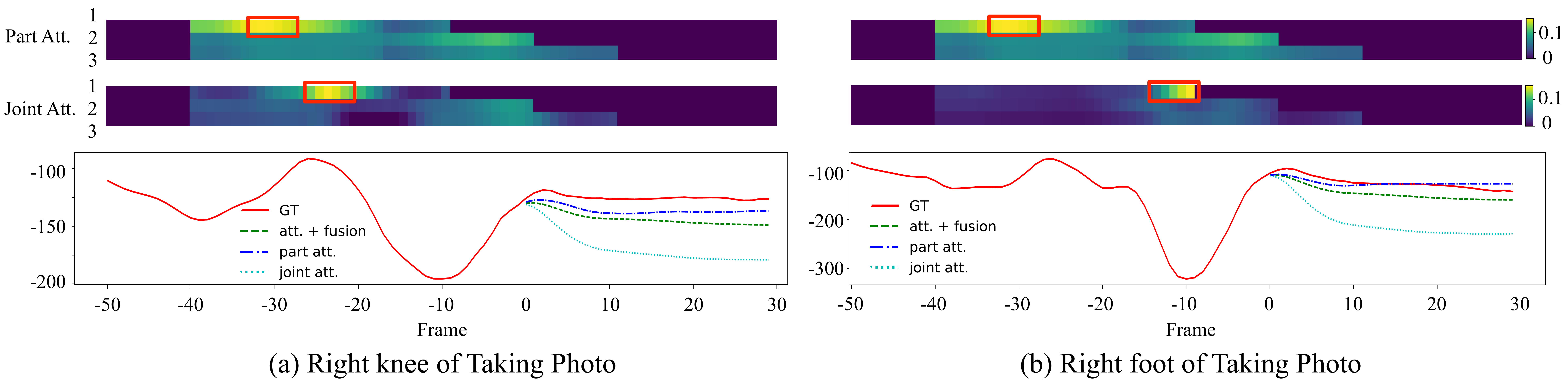}
      \caption{Visualization of attention maps and joint coordinate trajectories of different motion attentions for ``Taking Photo'' on H3.6M. At the first prediction step, as highlighted in red box, part motion attention, which generates one attention map for both joints by treating them as one body part, attends to historical motions that better reflect the current context.}
      \label{fig:h36-att-photo}
\end{figure*}
\begin{table*}[!ht]
\centering
\resizebox{0.8\textwidth}{!}{%
\begin{tabular}{c|cccccccc}
 & Walking & Eating & Smoking & Discussion & Directions & Greeting & Phoning & Posing \\\hline
Pose Motion Att.~\citep{mao2020history} & 58.1 & 75.7 & 69.5 & 119.8 & 106.5 & 138.8 & 105.0 & 178.2\\
Motion Att. + Post-fusion & \textbf{57.1} & \textbf{73.7} & \textbf{68.7} & \textbf{117.5} & \textbf{105.7} & \textbf{136.7} & \textbf{104.6} & \textbf{172.9}\\\hline\hline
 & Purchases & Sitting & SittingDown & TakingPhoto & Waiting & WalkingDog & WalkingTogether & Average \\\hline
Pose Motion Att.~\citep{mao2020history} & 134.2 & 115.9 & 143.6 & 115.9 & 108.2 & 146.9 & 64.9 & 112.1\\
Motion Att. + Post-fusion & \textbf{133.1} & \textbf{115.0} & \textbf{141.8} & \textbf{115.2} & \textbf{105.1} & \textbf{141.4} & \textbf{63.2} & \textbf{110.1}\\\hline
\end{tabular}%
}
\caption{Per-action 3D error at 1s on H3.6M.}
\label{tab:h36-action}
\end{table*}
\begin{table*}[!ht]
\centering
\resizebox{\textwidth}{!}{%
\begin{tabular}{c|cccccccccccccccccccccc}
 & R\_Knee (2) & R\_Ankle (3) & R\_Foot (4) & R\_Toe (5) & L\_Knee (7) & L\_Ankle (8) & L\_Foot (9) & L\_Toe (10) & Spine (12) & Neck (13) & Head (14) \\\hline
Pose Motion Att.~\citep{mao2020history} & 63.9 & 130.1 & 137.3 & 139.2 & 69.0 & 130.0 & 137.6 & 139.6 & 33.6 & 71.4 & 90.7 \\
Motion Att. + Post-fusion & \textbf{62.5} & \textbf{127.5} & \textbf{134.5} & \textbf{136.3} & \textbf{67.3} & \textbf{126.5} & \textbf{133.3} & \textbf{135.1} & \textbf{33.2} & \textbf{70.3} & \textbf{89.0} \\\hline\hline
 & Head Top (15) & L\_Shoulder (17) & L\_Elbow (18) & L\_Wrist (19) & L\_Site (21) & L\_Wrist End (22) & R\_Shoulder (25) & R\_Elbow (26) & R\_Wrist (27) & R\_Site (29) & R\_Wrist End (30) \\\hline
Pose Motion Att.~\citep{mao2020history} & 98.9 & 69.4 & 110.4 & 152.8 & 156.7 & 184.3 & 69.4 & 126.3 & 178.7 & 179.9 & 229.4 \\
Motion Att. + Post-fusion & \textbf{96.8} & \textbf{68.4} & \textbf{108.9} & \textbf{150.5} & \textbf{154.3} & \textbf{181.4} & \textbf{68.2} & \textbf{125.0} & \textbf{176.2} & \textbf{177.3} & \textbf{226.0}\\\hline
\end{tabular}
}
\caption{Per-joint 3D error at 1s on H3.6M. The numbers after the joint names are the joint index defined in H3.6M dataset. The index starts from $0$. Note that, we eliminate the joints that are fixed such as the ``Hip (0)''. For joints that share same 3D location, we only keep one of them in the table. For example, the 13th, 16th and 24th joints share the same 3D location, we thus only show the results on the 13th one.}
\label{tab:h36-joint}
\end{table*}

On one hand, modeling attention at a global level is not effective for motions whose local movements are not synchronized. Specifically, when the motion patterns of different body parts/joints are different, computing attention for them separately is more effective than using one shared attention. Such out of sync motions are common in non-periodical actions and sometimes even occur in periodical ones, such as the one shown in Figure~\ref{fig:h36-att-walking}. Specifically, in Figure~\ref{fig:h36-att-walking}, we compare the attention maps generated by pose motion attention (first row), joint motion attention (second row) and the predicted trajectories (third row) of two different joints (right hand and right foot) in a ``Walking'' sequence. As the motion patterns of the foot joint and hand joint are not synchronized, pose motion attention correctly captures the repeated pattern of the foot while attending to the wrong area for the hand. By contrast, joint motion attention, which generates two attention maps for these two joints, attends to the most relevant historical motions for both joints and leads to a better prediction.

\begin{table}[!ht]
\centering
\resizebox{\linewidth}{!}{%
\begin{tabular}{c|cccc|cccc}
\multirow{2}{*}{$\sigma$ (mm)} &\multicolumn{4}{c|}{Short-term} & \multicolumn{4}{c}{Long-term} \\ 
 & 80 & 160 & 320 & 400 & 560 & 720 & 880 & 1000 \\\hline
0 & 10.2 & 22.2 & 46.3 & 57.3 & 75.9 & 90.4 & 102.5 & 110.1 \\
2 & 13.1 & 26.0 & 50.3 & 61.0 & 78.8 & 92.6 & 104.3 & 111.6 \\
4 & 16.6 & 29.9 & 54.1 & 64.5 & 81.4 & 94.6 & 105.8 & 112.9 \\
6 & 19.8 & 33.4 & 57.3 & 67.4 & 83.9 & 96.6 & 107.6 & 114.5 \\
8 & 22.9 & 36.3 & 60.0 & 69.9 & 85.8 & 98.0 & 108.8 & 115.5 \\
10 & 26.0 & 39.6 & 62.7 & 72.4 & 87.9 & 99.8 & 110.4 & 117.3\\\hline
\end{tabular}%
}
\caption{Short \& long-term prediction of 3D joint positions on H3.6M with different levels of observation noise. The first column indicates the standard deviation (in millimeter) of the Gaussian noise added to the historical sequences. $\sigma=0$ means that no noise was added. }
\label{tab:h36-noise}
\end{table}
On the other hand, relying purely on local-level attention is not always optimal. Since local-level attention is computed from only the history of local movements, it may attend to sub-optimal areas in history, where different local body parts/joints have no or multiple similar motion patterns. Global-level attention helps to disambiguate the motion in such situations. We provide one example of this in Figure~\ref{fig:h36-att-photo}, where we show the attention maps and the trajectories predicted with part motion attention and joint motion attention for a ``Taking Photo'' sequence. Given the historical motion of individual joints only, joint motion attention wrongly attends to the area where a sharp motion in the negative direction occurs. By contrast, by leveraging information about complete body parts, part motion attention finds the historical motion that best reflects the current context.

As to quantitative results, we will provide an ablation study on fusing different motion  attention in section~\ref{sec:abla}. Here, we would like to emphasize that our multi-level motion attention fusion improves the motion prediction performance over pose motion attention only consistently for all actions. As shown in Table~\ref{tab:h36-action}, these improvements vary for actions of different natures. For instance, motion attention at the full pose level is sufficient to capture the motion patterns of periodical actions, such as ``Walking''; in such cases, the improvement obtained by our multi-level motion attention fusion model is indeed relatively small. By contrast, for other actions, such as ``Posing'' and ``Walking Dog'', fusing multi-level motion information yields significant improvements, of up to \textit{5 mm}, as evidenced by results in Table~\ref{tab:h36-action}.This is due to the fact that, in such actions, the repetitive motion patterns do not involve the full body but only body parts/joints.
    
To better understand the error distribution for each joint, we further show the 3D error for each joint separately after 1 second of prediction on H3.6M in Table~\ref{tab:h36-joint}. Our ``Motion Att. + Post-fusion'' consistently improves the performance on all joints. For some joints, such as ``Left Foot'', the improvements go up to \textit{4 mm}.

\begin{figure}[!t]
    \centering
      \includegraphics[width=0.8\linewidth]{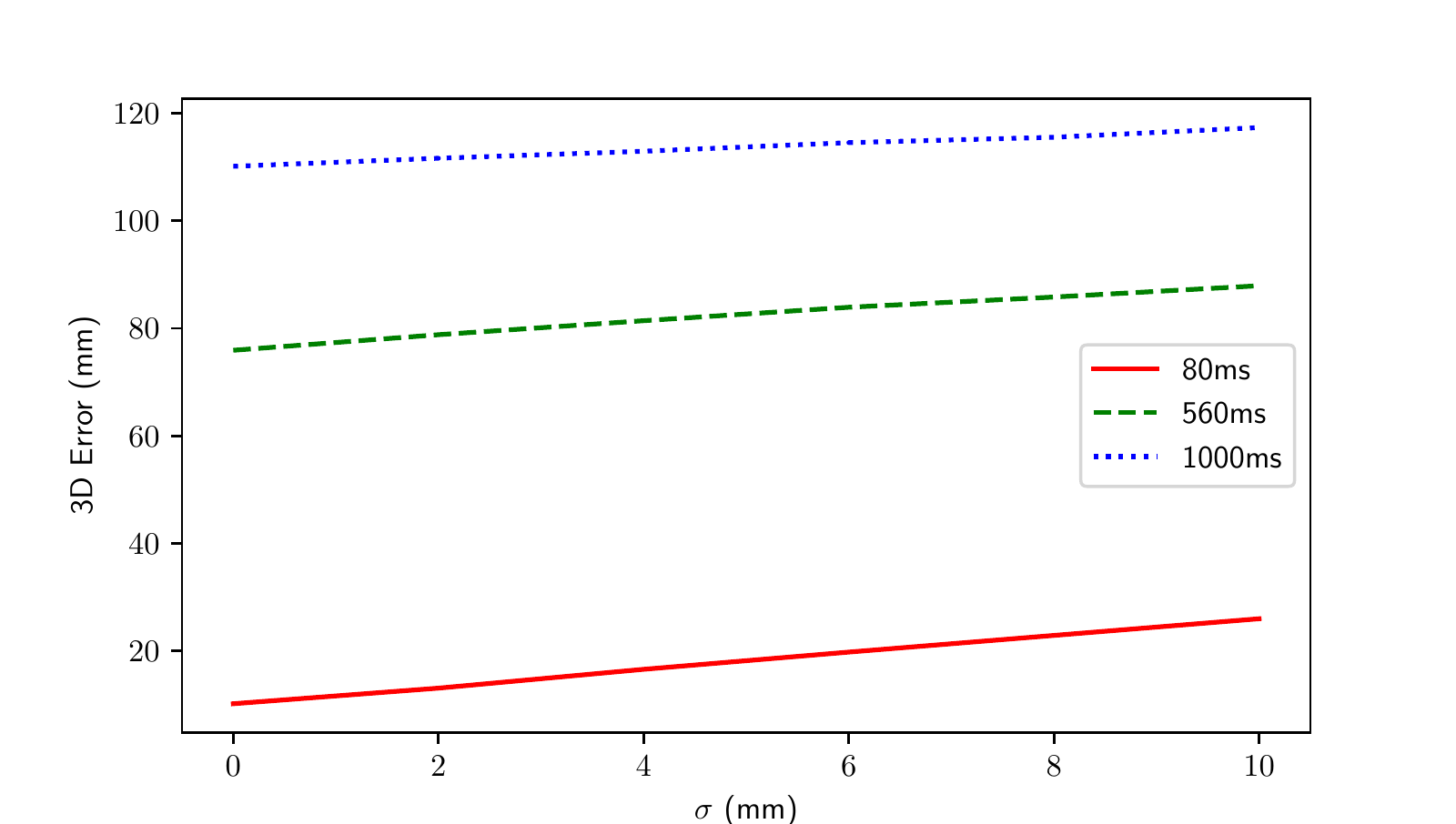}
      \caption{3D error vs noise levels.}
      \label{fig:h36-noise}
\end{figure}

\noindent{{\bf Influence of noisy history.}} We further study our model's ability of handling noisy history. In Table~\ref{tab:h36-noise}, we provide the results of our model obtained using observations corrupted by different levels of noise. Specifically, given the pretrained model, we added Gaussian noise ($\{\mathcal{N}(0,\,\sigma^{2})\}_{\sigma=\{0,2,4,6,8,10\}}$) to all joint coordinates of each frame in the history. As further shown in Figure~\ref{fig:h36-noise}, the 3D error grows linearly with the noise level ($\sigma$).

\begin{figure*}[!t]
    \centering
    \begin{tabular}{ccc}
    \includegraphics[width=0.5\linewidth]{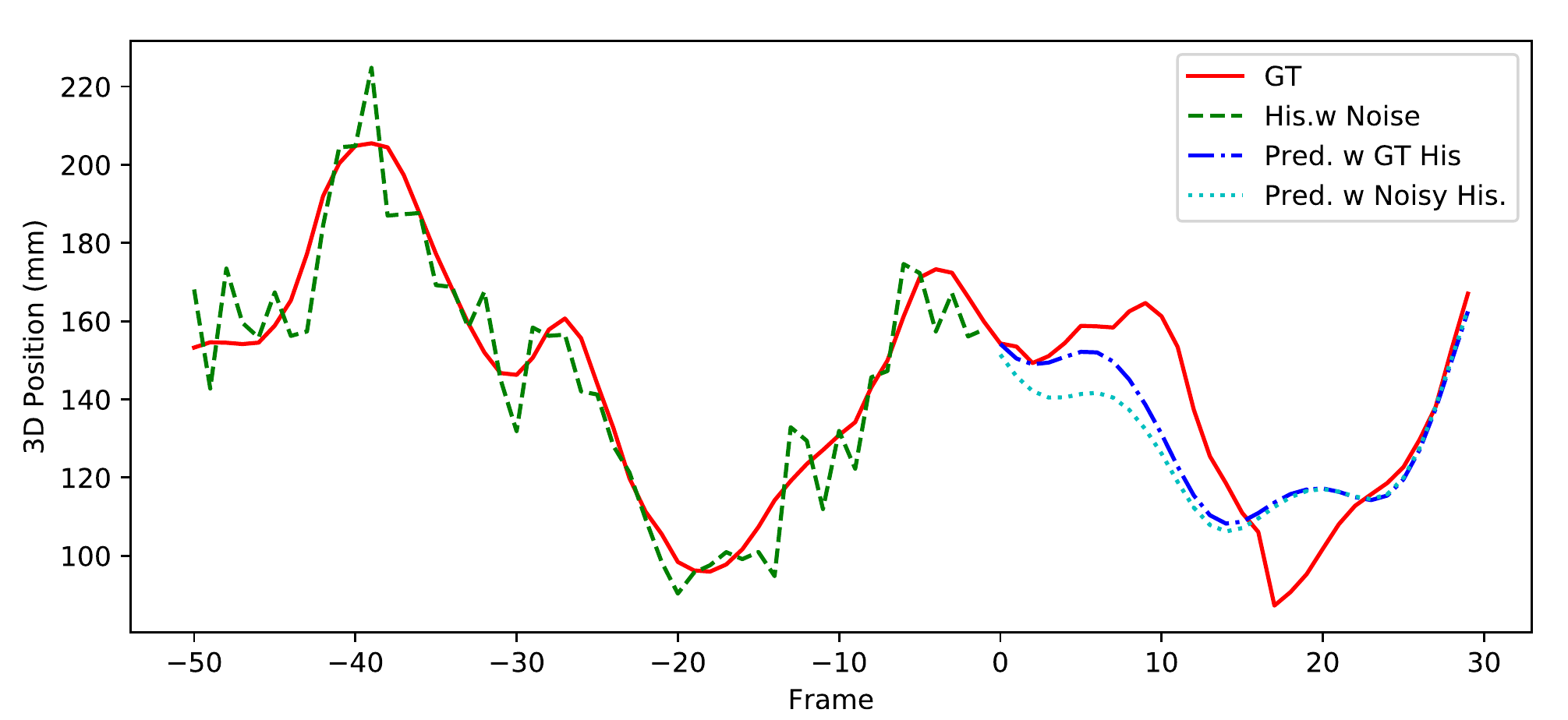} &
    \includegraphics[width=0.5\linewidth]{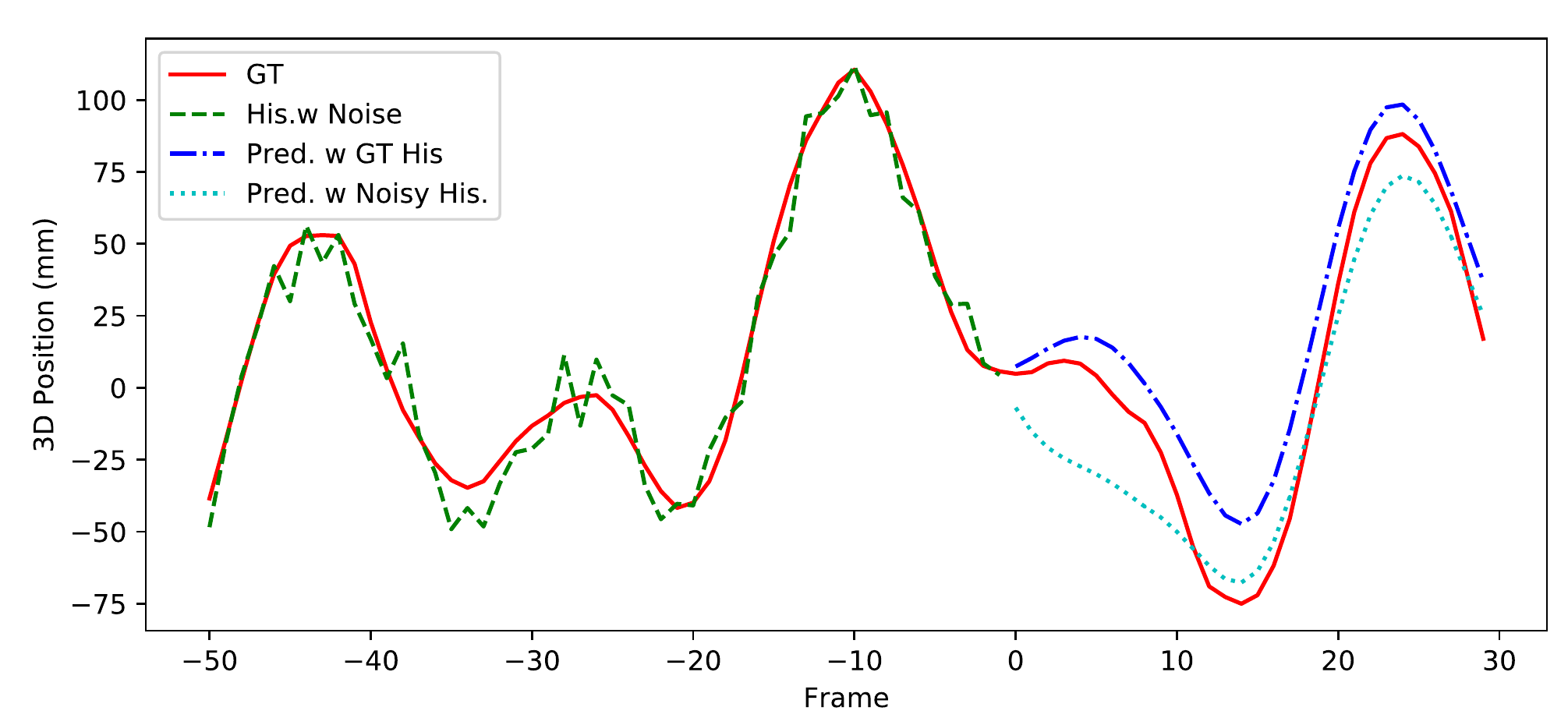}\\
    \end{tabular}
    \caption{Influence of jitter in the motion history. We created jitter by corrupting each historical pose with Gaussian noise~$\mathcal{N}(0,10)$. Our model still produces smooth future motions that are close to those predicted when using GT history.}
    \label{fig:jiter-his}
\end{figure*}

We further analyze the influence of jitter in Figure~\ref{fig:jiter-his}. Jitter was created by corrupting each historical pose with Gaussian noise~$\mathcal{N}(0,10)$. Our model is robust to such jitter and produces smooth future motions that are close to the ones predicted with the ground-truth history. 
This is because, instead of performing frame by frame prediction as in~\citep{Martinez_2017_CVPR}, our model generates a temporal encoding (DCT) of the sequence, which encourages global smoothness.

\begin{table}[!t]
\centering
\resizebox{0.5\textwidth}{!}{%
\begin{tabular}{ccccccccc}
milliseconds   & 80   & 160  & 320  & 400  & 560 & 720 & 880 & 1000 \\\hline
Concat. & 10.7 & 23.3 & 49.0 & 60.6 & 79.6 & 94.1 & 106.2 & 113.6 \\
Pre-fusion & 10.7 & 23.0 & 47.9 & 59.1 & 77.7 & 92.0 & 104.2 & 111.9 \\
Post-fusion & \textbf{10.2} & \textbf{22.2} & \textbf{46.3} & \textbf{57.3} & \textbf{75.9} & \textbf{90.4} & \textbf{102.5} & \textbf{110.1} \\\hline\hline
Concat. & 0.32 & 0.56 & 0.93 & 1.07 & 1.28 & 1.45 & 1.59 & 1.68 \\
Pre-fusion & \textbf{0.31} & \textbf{0.55} & \textbf{0.90} & 1.04 & \textbf{1.25} & 1.42 & 1.56 & \textbf{1.64} \\
Post-fusion & \textbf{0.31} & \textbf{0.55} & \textbf{0.90} & \textbf{1.03} & \textbf{1.25} & \textbf{1.40} & \textbf{1.55} & \textbf{1.64}
\\\hline
\end{tabular}
}
\caption{Comparison of different fusion strategies. ``Concat." corresponds to concatenating the outputs of all motion attention models, as shown in Fig.~\ref{fig:sele_model}(a).}
\label{tab:selec_stra}
\end{table}

\subsection{Ablation Study}~\label{sec:abla}
To further evaluate our fusion model, below, we first compare the performance of the different fusion strategies introduced in Section~\ref{sec:selec_model}. We then investigate the performance of fusing among different motion attention models.

\noindent{\bf Fusion strategies.} 
In Table~\ref{tab:selec_stra}, we compare the performance of three different fusion strategies. Post-fusion provides the best performance for both 3D joint positions and joint angles prediction.

\noindent{\bf Ablation on post-fusion.} In Table~\ref{tab:post_selec}, we evaluate the influence of fusing among the outputs of different motion attention models. For 3D joint position representation, the best results are obtained by fusing from all three motion attention models. By contrast, for joint angle representation, fusing among pose motion attention and part motion attention performs best. 

This is mainly due to the bias on training set. In particular, as shown in Table~\ref{tab:train_bias}, for joint angle representation, joint motion attention performs better than others on $59.3\%$ of the motion sequences in training set. After training on the biased training set, the fusion model tends to focus on predictions from joint motion attention model at all cases which leads to a inferior performance on the unbiased test set.

\begin{table*}[!t]
\centering
\resizebox{0.85\textwidth}{!}{%
\begin{tabular}{cccc|ccccccccc}
Pose Att. & Part Att. & Joint Att. & Post-fusion   & 80ms   & 160ms  & 320ms  & 400ms  & 560ms & 720ms & 880ms & 1000ms \\\hline
\checkmark & & & & 10.4 & 22.6 & 47.1 & 58.3 & 77.3 & 91.8 & 104.1 & 112.1 \\
 & \checkmark & & & 10.5 & 22.7 & 47.6 & 58.8 & 77.5 & 92.0 & 104.3 & 112.0 \\
 & & \checkmark & & 10.5 & 22.9 & 48.1 & 59.6 & 78.4 & 92.8 & 104.9 & 112.3 \\
\checkmark & \checkmark & & \checkmark &  10.4 & 22.6 & 47.1 & 58.3 & 77.3 & 91.8 & 104.1 & 112.1 \\
\checkmark & & \checkmark & \checkmark & 10.3 & 22.3 & 46.5 & 57.6 & 76.3 & 90.7 & 102.9 & 110.5 \\
& \checkmark & \checkmark & \checkmark & 10.5 & 22.9 & 48.1 & 59.6 & 78.4 & 92.8 & 104.9 & 112.3 \\
\checkmark & \checkmark & \checkmark & \checkmark & \textbf{10.2} & \textbf{22.2} & \textbf{46.3} & \textbf{57.3} & \textbf{75.9} & \textbf{90.4} & \textbf{102.5} & \textbf{110.1} \\\hline\hline
\checkmark & & & & \textbf{0.31} & \textbf{0.55} & \textbf{0.90} & 1.04 & \textbf{1.25} & 1.42 & 1.56 & 1.65 \\
 & \checkmark & & & 0.32 & 0.56 & 0.92 & 1.05 & 1.26 & 1.42 & 1.57 & 1.65 \\
 & & \checkmark & & \textbf{0.31} & \textbf{0.55} & 0.92 & 1.05 & 1.27 & 1.43 & 1.57 & 1.66 \\
\checkmark & \checkmark & & \checkmark & \textbf{0.31} & \textbf{0.55} & \textbf{0.90} & \textbf{1.03} & \textbf{1.25} & \textbf{1.40} & \textbf{1.55} & \textbf{1.64} \\
\checkmark & & \checkmark & \checkmark & \textbf{0.31} & \textbf{0.55} & 0.92 & 1.05 & 1.27 & 1.43 & 1.57 & 1.66 \\
& \checkmark & \checkmark & \checkmark & \textbf{0.31} & \textbf{0.55} & 0.92 & 1.05 & 1.27 & 1.43 & 1.57 & 1.66 \\
\checkmark & \checkmark & \checkmark & \checkmark & \textbf{0.31} & \textbf{0.55} & 0.91 & 1.04 & \textbf{1.25} & 1.41 & 1.56 & \textbf{1.64} \\\hline
\end{tabular}
}
\caption{Ablation on post-fusion strategy. We compare the average 3D joint position error (upper) and joint angle error (bottom) on H3.6M. For 3d joint position, best performance is obtained by fusing among all 3 motion attention models. For joint angle, fusing pose motion attention and part motion attention performs the best.}
\label{tab:post_selec}
\end{table*}
\begin{table*}[!t]
\centering
\resizebox{0.75\textwidth}{!}{%
\begin{tabular}{cc|cccccccc|c}
 & & 80ms   & 160ms  & 320ms  & 400ms  & 560ms & 720ms & 880ms & 1000ms & percent\\\hline
\multirow{3}{*}{Test set}
&Pose Att. & \textbf{0.31} & \textbf{0.55} & \textbf{0.90} & \textbf{1.04} & \textbf{1.25} & \textbf{1.42} & \textbf{1.56} & \textbf{1.65} & 33.5\%\\
&Part Att. & 0.32 & 0.56 & 0.92 & 1.05 & 1.26 & 1.42 & 1.57 & \textbf{1.65} & 31.1\% \\
&Joint Att. & \textbf{0.31} & \textbf{0.55} & 0.92 & 1.05 & 1.27 & 1.43 & 1.57 & 1.66 & 35.3\%\\\hline
\multirow{3}{*}{Training set}
& Pose Att. & 0.29 & 0.50 & 0.81 & 0.94 & 1.17 & 1.34 & 1.49 & 1.58 & 21.5\% \\
& Part Att. & 0.29 & 0.50 & 0.81 & 0.94 & 1.17 & 1.33 & 1.48 & 1.57 & 19.2\% \\
& Joint Att. & \textbf{0.28} & \textbf{0.48} & \textbf{0.77} & \textbf{0.89} & \textbf{1.12} & \textbf{1.29} & \textbf{1.44} & \textbf{1.54} & 59.3\%\\\hline
\end{tabular}
}
\caption{Performance bias on training set comparing to test set of 3 different attention models. We show the average angle error on test set (top) and training set (bottom) of H3.6M. Besides, the last column demonstrates the percentage of each type of motion attention model outperforms the others among all sequences. For example, the joint attention model outperforms others on $59.3\%$ training samples, leading to a consistent better performance across all time horizons. However, all 3 attention models perform comparable to each other on test set.}
\label{tab:train_bias}
\end{table*}
\noindent{\bf GCNs vs fully-connected networks.} Finally, we evaluate the importance of using GCNs vs fully-connected networks and of learning the connectivity in the GCN instead of using a pre-defined adjacency matrix based on the kinematic tree. The results of these experiments, provided in Table~\ref{table-ablation-fullyconnect}, demonstrate the benefits of both using GCNs and learning the corresponding graph structure. Altogether, this ablation study indicates the importance of both aspects of our contribution: Using the DCT to model temporal information and learning the connectivity in GCNs to model spatial structure.
\begin{table*}[!t]
\centering
\resizebox{\linewidth}{!}{%
\begin{tabular}{c|cccc|cccc|cccc|cccc|cccc}
  & \multicolumn{4}{c}{Walking} & \multicolumn{4}{c}{Eating} & \multicolumn{4}{c}{Smoking} & \multicolumn{4}{c}{Discussion}&\multicolumn{4}{c}{Average} \\
  & 80 & 160 & 320 & 400 & 80 & 160 & 320 & 400 & 80 & 160 & 320 & 400 & 80 & 160 & 320 & 400& 80 & 160 & 320 & 400 \\ \hline
  Fully-connected network & 0.20 & 0.34 & 0.54 & 0.61 & 0.18 & 0.31 & 0.53 & 0.66 & 0.22 & 0.43 & \textbf{0.85} & 0.83 & 0.28 & 0.64 & 0.87 & 0.93 & 0.22 & 0.43 & 0.70 & 0.76\\
 with pre-defined connectivity &  0.25 & 0.46 & 0.70 & 0.8 & 0.23 & 0.41 & 0.68 & 0.83 & 0.24 & 0.46 & 0.93 & 0.91 & 0.27 & 0.62 & 0.89 & 0.97 & 0.25 & 0.49 & 0.80 & 0.88\\
 with learnable connectivity & \textbf{0.18} & \textbf{0.31} & \textbf{0.49} & \textbf{0.56} & \textbf{0.16} & \textbf{0.29} & \textbf{0.50} & \textbf{0.62} & \textbf{0.22} & \textbf{0.41} & 0.86 & \textbf{0.80} & \textbf{0.20} & \textbf{0.51} & \textbf{0.77} & \textbf{0.85} & \textbf{0.19} & \textbf{0.38} & \textbf{0.66} & \textbf{0.71}\\
 \hline\hline
  Fully-connected network & 11.2 & 18.6 & 33.5 & 38.8 & 9.0 & 18.8 & \textbf{39.0} & 48.0 & 8.5 & 15.4 & 26.3 & 31.4 & 12.2 & 26.0 & 46.3 & 53.0 & 10.2 & 19.7 & 36.3 & 42.8\\
 with pre-defined connectivity & 25.6 & 44.6 & 80.3 & 96.8 & 16.3 & 31.9 & 62.4 & 78.8 & 11.6 & 21.4 & 34.6 & 38.6 & 20.7 & 38.7 & 62.5 & 69.9 & 18.5 & 34.1 & 59.9 & 71.0 \\
 with learnable connectivity & \textbf{8.9} & \textbf{15.7} & \textbf{29.2} & \textbf{33.4} & \textbf{8.8} & \textbf{18.9} & 39.4 & \textbf{47.2} & \textbf{7.8} & \textbf{14.9} & \textbf{25.3} & \textbf{28.7} & \textbf{9.8} & \textbf{22.1} & \textbf{39.6} & \textbf{44.1} & \textbf{8.8} & \textbf{17.9} & \textbf{33.4} & \textbf{38.4}\\
 \hline
\end{tabular}
}
\caption{Influence of GCNs and of learning the graph connectivity. Top: angle error on 8 sequences per action; Bottom: 3D error on 8 sequences per action. Note that GCNs with a pre-defined connectivity yield much higher errors than learning this connectivity as we do. Here, we reused the results from~\citep{mao2019learning}.}
\label{table-ablation-fullyconnect}
\end{table*}

\section{Conclusion}
In this paper, we have introduced an attention-based motion prediction approach that selectively exploits historical information according to the similarity between the current motion context and the sub-sequences in the past. This has led to a predictor equipped with a~\emph{motion attention} model that can effectively make use of historical motions, even when they are far in the past. Furthermore, we have studied the use of motion attention at different levels, full body, body parts, joints, and shown that combining these different attention levels led to better performance. Our approach achieves state-of-the-art performance on the commonly-used motion prediction benchmarks and on recently-published datasets. Moreover, our experiments have demonstrated that our network generalizes to previously-unseen datasets without re-training or fine-tuning, and can handle longer history than that it was trained with to further boost performance on non-periodical motions with repeated history. In the future, we will investigate the combination of our approach with the progressive prediction strategy of~\cite{cai2020learning}.

\section*{Acknowledgements}
This research was supported in part by the Australia Research Council DECRA Fellowship (DE180100628) and ARC Discovery Grant (DP200102274). The authors would like to thank NVIDIA for the donated GPU (Titan V).


%
%

\bibliographystyle{spbasic}      
\bibliography{egbib}   


\clearpage
\setcounter{figure}{0} 
\setcounter{table}{0} 
\setcounter{section}{0}

\newcommand*{\dictchar}[1]{
	\clearpage
	\twocolumn[
	\centerline{\parbox[c][3cm][c]{15cm}{
			\centering
			\selectfont
			{#1}}}]
}

\bigskip
\dictchar{\noindent \Large \bf {Multi-level Motion Attention for Human Motion Prediction\\-----Supplementary Material-----}}

\section{Datasets}
Below we provide more details about the datasets used in our experiments.

\noindent{\textbf{Human3.6M.}} As in~\citep{mao2019learning}, we use the skeleton of the subject 1 (S1) of Human3.6M as standard skeleton to compute the 3D joint coordinates from the joint angle representation. After removing the global rotation, translation and constant angles or 3D
coordinates of each human pose, this leaves us with a 48 dimensional vector and a 66 dimensional vector for human pose in angle representation and 3D position, respectively. As in~\citep{mao2019learning,LiZLL18,Martinez_2017_CVPR}, the rotation angles are represented as exponential maps. During training, we set aside subject 11 (S11) as our validation set to choose the model that achieves the best performance across all future frames, and the remaining 5 subjects (S1,S6,S7,S8,S9) are used as training set.

\noindent{\textbf{AMASS \& 3DPW.}} The human skeleton in AMASS and 3DPW is defined by a shape vector. In our experiment, we obtain the 3D joint positions by applying forward kinematic on the skeleton derived from the shape vector of the CMU dataset. As specified in the main paper, we evaluate the model on BMLrub and 3DPW. Each video sequence is first downsampled to 25 frames per second, and evaluate on sub-sequences of length $M+T$ that start from every $5^{th}$ frame of each video sequence.
\section{Implementation Details}
We implemented our network in Pytorch~\citep{paszke2017automatic} and trained it using the ADAM optimizer~\citep{kingma2014adam}. We use a learning rate of $0.0005$ with a decay at every epoch so as to make the learning rate be $0.00005$ at the $50^{th}$ epoch. We train our model for $50$ epochs with a batch size of 32 for H3.6M and 128 for AMASS. One forward and backward pass takes 32ms for H3.6M and 45ms for AMASS on an NVIDIA Titan V GPU.

For post-fusion strategy, we first train the three different level of attention model as well as their predictors for 50 epochs. After that, we fix the attention models and predictors and use the output of the predictors to train the fusion model for another 20 epochs.
\section{Generating Long Future for Periodical Motions}
For periodical motions, such as ``Walking", our approach can generate very long futures (up to 16 seconds). As shown in the supplementary video, such future predictions are hard to distinguish from the ground truth even for humans.
\section{Additional Results on AMASS}
\begin{figure*}[t]
	\centering
	\begin{tabular}{c}
		\includegraphics[width=\linewidth]{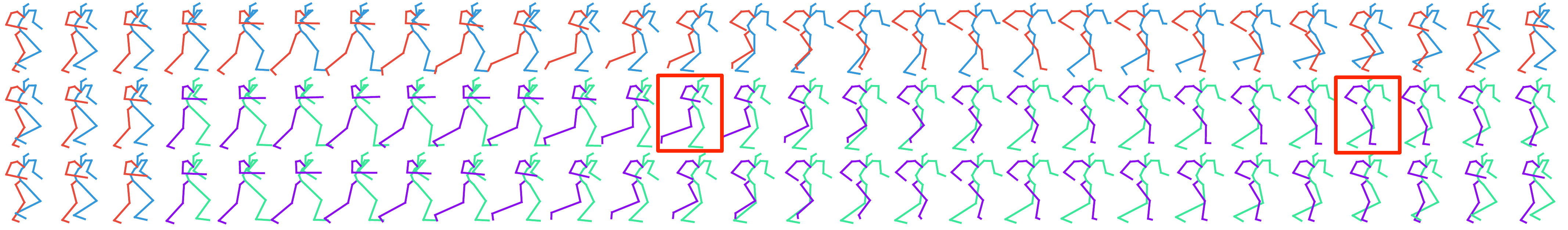} \\
		(a) Jogging\\
		\includegraphics[width=\linewidth]{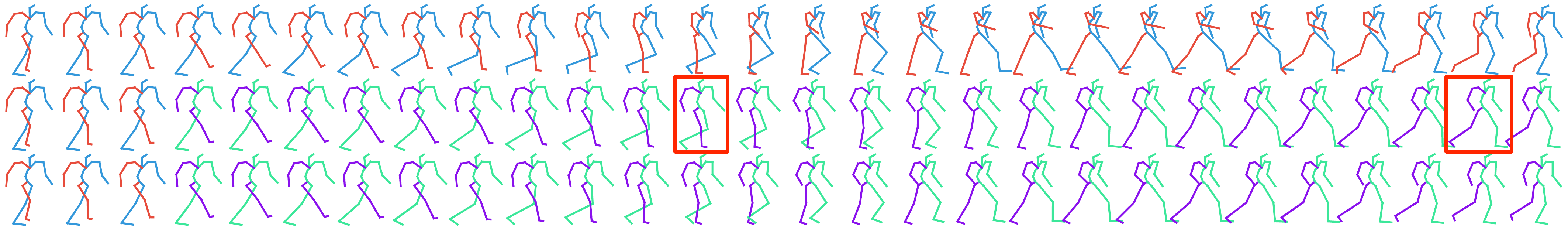} \\
		(b) Walking\\
		\includegraphics[width=\linewidth]{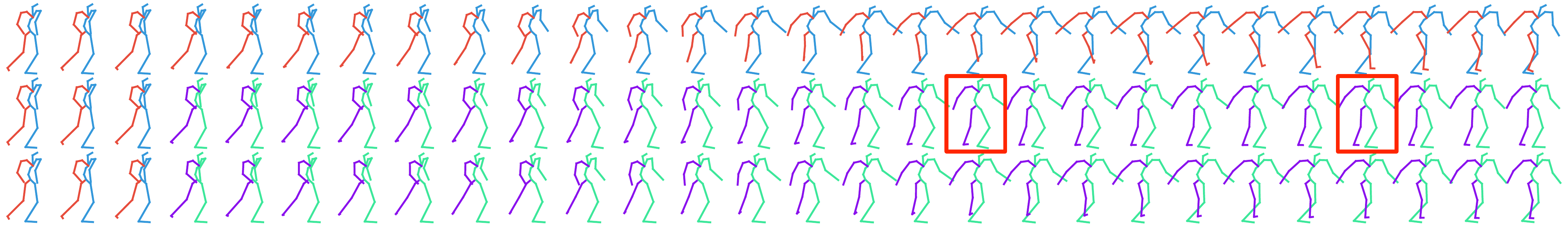} \\
		(c) Stretching
	\end{tabular}
	\caption{Qualitative comparison on the BMLrub dataset. From top to bottom, we show the ground-truth motion, the prediction results of LTD~\citep{mao2019learning} and of our approach on 3D position. The observed poses are shown as blue and red skeletons and the predictions in green and purple. As highlighted by the red boxes, our predictions better match the ground truth, in particular for the legs.}
	\label{fig:vis_amass}
\end{figure*}

In Fig.~\ref{fig:vis_amass}, we compare the results of LTD~\citep{mao2019learning} and of our approach on the BMLrub dataset. Our results better match the ground truth.
\section{Motion Attention vs. Frame-wise Attention}
To further investigate the influence of \emph{motion} attention, where the attention on the history sub-sequences $\{\textbf{X}_{i:i+M+T-1}\}_{i=1}^{N-M-T+1}$ is a function of the first $M$ poses of every sub-sequence $\{\textbf{X}_{i:i+M-1}\}_{i=1}^{N-M-T+1}$ (keys) and the last observed $M$ poses $\textbf{X}_{N-M+1:N}$ (query), we replace the keys and query with the last frame of each sub-sequence. That is, we use $\{\textbf{X}_{i+M-1}\}_{i=1}^{N-M-T+1}$ as keys and $\textbf{X}_{N}$ as query. We refer to the resulting method as \emph{Frame-wise Attention}. As shown in Table~\ref{tab:ablation}, motion attention outperforms frame-wise attention by a large margin. As discussed in the main paper, this is due to frame-wise attention not considering the direction of the motion, leading to ambiguities.

\begin{table*}[!ht]
\centering
\resizebox{0.8\textwidth}{!}{%
	\begin{tabular}{ccccccccc}
		milliseconds   & 80   & 160  & 320  & 400  & 560 & 720 & 880 & 1000 \\\hline
		Frame-wise Attention & 24.0 & 44.5 & 76.1 & 88.3 & 107.5 & 121.7 & 131.7 & 136.7\\
		Motion Attention & \textbf{10.8} & \textbf{23.9} & \textbf{49.4} & \textbf{60.7} & \textbf{77.3} & \textbf{92.0} & \textbf{104.4} & \textbf{112.4}\\\hline
	\end{tabular}
}
\caption{Comparison of frame-wise attention and with our motion attention.}
\label{tab:ablation}
\end{table*}

\end{document}